\RequirePackage{fix-cm}     

\documentclass[a4paper,fleqn]{cas-sc}

\def\tsc#1{\csdef{#1}{\textsc{\lowercase{#1}}\xspace}}
\tsc{WGM}
\tsc{QE}

\usepackage{amsmath,amsfonts}
\usepackage{algorithmic}
\usepackage{algorithm}
\usepackage[caption=false,font=normalsize,labelfont=sf,textfont=sf]{subfig}
\usepackage{textcomp}
\usepackage{url}
\usepackage{verbatim}
\usepackage{graphicx}
\usepackage[space]{cite}
\usepackage{xcolor}

\usepackage{orcidlink}   
\usepackage{etoolbox}
\AtBeginDocument{
    \renewcommand{\cite}[1]{\textcolor{blue}{[\citenum{#1}]}}
}
   
\usepackage{csquotes}       
\usepackage[capitalise]{cleveref}		
\crefname{figure}{Fig.}{Figs.}  
\Crefname{figure}{Fig.}{Figs.}  
\crefname{equation}{Eq.}{Eqs.}
\Crefname{equation}{Eq.}{Eqs.}
\crefrangeformat{equation}
{Eqs.~(#3#1#4) - (#5#2#6)}

\Crefrangeformat{equation}
{Eqs.~(#3#1#4) - (#5#2#6)}

\usepackage{color,soul}         
\usepackage{amssymb}   
\usepackage{pifont}

\usepackage{array}
\usepackage{booktabs}
\usepackage{vcell}
\usepackage{tabularx} 
\usepackage{fontsize} 
\usepackage{makecell} 
\usepackage{multirow}  
\usepackage{adjustbox} 
\definecolor{mydarkgreen}{RGB}{0,200,0}

\renewcommand\tabularxcolumn[1]{m{#1}}
\usepackage{tabularray}
\usepackage{acro}[=v2]    
\usepackage{enumitem}
\usepackage{mfirstuc}   

\setlist{nolistsep}    

\usepackage{tikz}
\usetikzlibrary{arrows.meta, shapes.geometric} 

\newcommand{\AuthorORCID}[1]{%
  \texorpdfstring{\raisebox{0.5ex}{\orcidlink{#1}}}{}%
}

\ExplSyntaxOn
\RenewDocumentCommand \printemails { }
{
  \group_begin:
  \renewcommand{\AuthorORCID}[1]{}
  \int_compare:nNnTF { \int_use:N \g_ead_int } > { 0 }
  {
    \begin{NoHyper}
    \tex_let:D \thefootnote \relax
    \footnotetext
    {
      \raggedright
      \bool_if:NTF \g_stm_nologo_bool
      { 
        \int_compare:nTF { \g_ead_int = 1 }
        { \textit{Email~address:\c_space_token} }
        { \textit{Email~addresses:\c_space_token} }
      }
      { \includegraphics[height=8pt]{thumbnails/cas-email.jpeg}\c_space_token } 
      \seq_use:Nn \g_stm_ead_seq { ;~ }
    }
    \end{NoHyper}
  }
  {  }
  \group_end:
}

\RenewDocumentCommand \printcornotes { }
{
  \begin{NoHyper}
  \seq_use:Nn \g_stm_cor_seq { \par }
  \end{NoHyper}
}
\ExplSyntaxOff

\begin{document}
\hypersetup{
  bookmarksopen=true,
  bookmarksopenlevel=2,
  bookmarksdepth=3
}
\def\floatpagepagefraction{1}
\def\textpagefraction{.001}

\shorttitle{An Agentic AI Framework with LLMs and CoT for UAV-Assisted Logistics Scheduling with MEC}    

\shortauthors{H. Zhang; D. Niyato; W. Zhang; X. Lou; M.Y.H. Low}  

\title [mode = title]{An Agentic AI Framework with Large Language Models and Chain-of-Thought for UAV-Assisted Logistics Scheduling with Mobile Edge Computing}  

\author[address1,address2]{Hanwen Zhang\AuthorORCID{0000-0002-6295-4753}}

\ead{hanwen001@e.ntu.edu.sg} \ead{hanwen.zhang@singaporetech.edu.sg}

\affiliation[address1]{
            addressline={College of Computing and Data Science, Nanyang Technological University}, 
            city={Singapore},
            postcode={639798}, 
            country={Singapore}}

\affiliation[address2]{
            addressline={Information and Communications Technology Cluster, Singapore Institute of Technology}, 
            city={Singapore},
            postcode={828608}, 
            country={Singapore}}

\author[address1]{Dusit Niyato\AuthorORCID{0000-0002-7442-7416}}
\ead{dniyato@ntu.edu.sg}

\author[address2]{Wei Zhang\AuthorORCID{0000-0002-2644-2582}}
\cormark[1]
\ead{wei.zhang@singaporetech.edu.sg}

\author[address2]{Xin Lou\AuthorORCID{0000-0001-8910-5666}}
\ead{lou.xin@singaporetech.edu.sg}

\author[address2]{Malcolm Yoke Hean Low\AuthorORCID{0000-0002-5371-0896}}
\ead{malcolm.low@singaporetech.edu.sg}

\cortext[1]{Corresponding author}

\begin{abstract}
In cloud manufacturing, unmanned aerial vehicles (UAVs) can support both product collection and mobile edge computing (MEC). This joint operation forms a hybrid scheduling problem, where physical logistics decisions are coupled with computational task scheduling. In this paper, UAVs collect finished products from manufacturing stations and transport them back to a central depot. Meanwhile, computational tasks generated by industrial sensor devices at these stations are processed locally, at UAVs, or offloaded via UAVs to the cloud. This coupling makes the problem challenging. A UAV can provide MEC services only during its service window at a station, so routing decisions directly determine when UAV-assisted offloading is available. Routing decisions also affect the UAV energy budget and the availability of onboard computing and communication resources for computational task execution under task deadline constraints. To address this, we propose an agentic-AI-assisted optimization framework with two components. First, we develop an agentic AI that combines large language models, retrieval-augmented generation, and chain-of-thought reasoning to translate user input into an interpretable mathematical formulation for the hybrid scheduling problem. Second, we design a hierarchical deep reinforcement learning approach based on proximal policy optimization (PPO), where the upper layer learns UAV routing and the lower layer optimizes per-slot task execution and resource allocation. Simulation results show that the proposed framework yields more consistent formulations, while the hierarchical PPO achieves full product collection in 99.6\% of the last 500 episodes and maintains a 100\% deadline satisfaction rate, with more stable performance than the advantage actor-critic approach. 
\end{abstract}


\maketitle

\section{Introduction}

Large language models (LLMs) can support manufacturing scheduling by interpreting natural-language requirements and organizing domain knowledge into mathematical formulation logic. This capability is important when task descriptions, operational constraints, and solver-oriented representations must be connected consistently. To address this challenge, the authors in \cite{Ref_CoT_Mfg3} developed an LLM-based multi-agent scheduling chain for the flexible job shop scheduling problem. In this framework, an improved decision-making process is used to analyze scheduling indicators and select suitable algorithms. The overall framework also supports dynamic scheduling and real-time rescheduling. More recently, \cite{Ref_CoT_Mfg4} proposed an LLM-empowered dynamic scheduling framework for intelligent hybrid flow shops. In this framework, structured prompt engineering is used to enrich state representation and guide action selection within a multi-agent deep reinforcement learning (DRL) framework. LLMs have also been used for combinatorial optimization \cite{CP1_GraphThought}, job shop scheduling \cite{JSP3_LLM}, and dynamic multi-agent manufacturing systems \cite{JSP2_LLM}.

This formulation-oriented role of LLMs is particularly relevant to cloud manufacturing (CMfg), where scheduling problems often couple physical logistics with computational services. CMfg enables service-oriented sharing and efficient allocation of geographically distributed resources, and coordinated decision-making across cyber-physical components is essential to achieve its value \cite{JSP_DRL18_R1}. With the development of Internet of Things (IoT) technologies, CMfg increasingly relies on cloud-edge-device collaboration. In this setting, industrial sensor devices (ISDs) continuously generate sensing data for production monitoring, equipment-status tracking, quality inspection, and anomaly detection. These data flows require timely transmission and processing to support reliable industrial IoT decisions. Computational tasks such as digital twins, real-time analytics, and predictive maintenance therefore support time-sensitive industrial operations \cite{JSP_DRL18_CMfg}. Physical logistics can be executed by automated agents, such as unmanned aerial vehicles (UAVs) \cite{TwoLayerDRL} or automated guided vehicles \cite{JSP_DRL17}, while computational workloads are processed at the device, edge, or cloud layers \cite{sun_UVA}. In this paper, we study an integrated scheduling problem that jointly optimizes UAV-assisted product collection and cross-layer task offloading from devices to UAVs and the cloud. In the considered system, a UAV provides more than product-collection mobility. When the UAV visits a station, it also creates a temporary wireless access opportunity for ISD-to-UAV uplink transmission and UAV-to-cloud backhaul forwarding. The resulting service window determines when UAV-assisted processing and cloud offloading through the UAV-assisted path are available for sensing-data tasks. Therefore, UAV routing directly affects wireless connectivity, edge-processing latency, and the reliability of deadline-sensitive industrial IoT services. The problem is challenging because UAV routes induce service windows for UAV-assisted MEC, while task execution consumes computing, communication, and energy resources under deadline constraints.

\textbf{Challenge I}: 
The mathematical modeling of combinatorial optimization problems in CMfg can be significantly more complex than that of traditional single-domain scheduling problems. Similar coupling complexity has also been observed in UAV-assisted MEC systems, where computational task offloading is jointly optimized with UAV trajectory, path-planning, or task-scheduling decisions \cite{Ref_hyb2}. It requires capturing how UAV routing induces time-varying service windows. These service windows, in turn, affect per-slot task offloading and compute/communication allocation under energy budgets, capacity limits, and task deadlines. At the same time, the model must balance conflicting goals, such as logistics efficiency and sufficient resource allocation for meeting task deadlines. These complexities lead to a complicated modeling process. As a result, the problem becomes challenging for engineers and researchers to model, formulate, and solve such complex problems. This issue is due to the fact that specific operational dynamics of joint logistics-computation scheduling CMfg have to be taken into account. 
\textbf{Challenge II}: 
Compared with traditional scheduling problems, hybrid task scheduling in CMfg introduces significantly higher solution complexity. The reason is that logistics- and computation-related decisions must be optimized jointly rather than handled in isolation. Moreover, as user demands and system conditions evolve over time, the resulting optimization problem may also change. Therefore, the key challenge is to solve these changing optimization problems efficiently. In particular, the scheduler must jointly determine UAV routing, offloading-destination selection, and compute/communication allocation under strict latency, resource, and communication constraints. Therefore, designing tractable DRL frameworks for hybrid task scenarios with sparse feedback and complex action representations is crucial for enhancing the efficiency of intelligent CMfg systems.

To formulate complex combinatorial problems in CMfg, we present an agentic AI framework to address \textbf{Challenge I}. The agent facilitates dialogue-driven mathematical model generation and refinement by combining LLM reasoning with retrieval-augmented generation (RAG) and chain-of-thought (CoT) reasoning. In particular, an LLM converts natural language descriptions into structured modeling logic. Then, RAG retrieves task-relevant modeling principles and generic formulation primitives from a curated domain-specific knowledge base. CoT further makes explicit the logic underlying variable definitions, constraint design, and objective construction. Hence, CoT can enhance the accuracy and transparency of the model. In CMfg settings where decisions couple UAV routing, offloading-destination selection, and compute/communication resource allocation, the proposed agentic AI helps UAV operators efficiently derive interpretable, domain-specific formulations.

To address \textbf{Challenge II}, we propose a hierarchical DRL approach to solve the coupled UAV-assisted collection and mobile edge computing (MEC) scheduling problem. Inspired by \cite{TwoLayerDRL}, the framework decomposes the original joint optimization into two tractable sequential decision layers. This decomposition substantially reduces the effective state/action space and retains the forward operational coupling from UAV routing to MEC scheduling through routing-induced information. Both layers are formulated as finite-horizon Markov decision processes (MDPs). They are solved using proximal policy optimization (PPO). This helps maintain stable policy updates under high-dimensional and constrained decision making. Specifically, the \emph{upper layer} (UAV routing) formulates the multi-UAV routing problem as a single-agent MDP. It jointly determines the manufacturing-station assignment for each UAV and the corresponding visiting order under energy, payload, and flight-distance constraints. Given the resulting routes, the \emph{lower layer} (computational task assignment) makes per-slot execution decisions---offloading destination (local/UAV/cloud via UAV) and compute/communication allocation---under capacity and deadline constraints. The two layers interact through routing-induced signals. The upper layer provides service windows and remaining UAV energy to the lower layer, which uses them to determine the feasibility of UAV-involved processing. 

This is the first work in the field of smart manufacturing that utilizes agentic AI, i.e., LLMs with RAG and CoT reasoning, to formulate hybrid combinatorial optimization problems in CMfg. The formulated problems are then solved using a hierarchical DRL approach. The contributions of this work are summarized as follows: 
\begin{itemize}
    \item We propose an agentic AI framework that integrates LLMs, RAG, and CoT reasoning to assist in formulating hybrid combinatorial problems in CMfg. This framework enables the decomposition of complex task requirements into structured optimization models, capturing the intricate dependencies among logistics and computational tasks. (For \textbf{Challenge I})
    \item We design an interactive problem formulation process that leverages RAG to retrieve domain-specific knowledge and CoT to enhance interpretability and traceability of the generated models. This process allows the agentic AI to reason step-by-step about task couplings, performance requirements, resource constraints, and scheduling objectives. Hence, the process enables accurate and adaptive model construction under cloud-edge-device collaboration. (For \textbf{ Challenge I})
    \item We develop a hierarchical DRL approach to solve the formulated hybrid scheduling problem. By decomposing the coupled routing-MEC decision making into an upper routing layer and a lower task-scheduling layer, the framework reduces the effective joint decision complexity while retaining route-to-scheduling coordination through routing-induced service availability. (For \textbf{Challenge II})
\end{itemize}

The remainder of this paper is organized as follows. \Cref{sec:Related_Work} reviews the relevant literature. \Cref{sec:SystemModel} then presents the system model for UAV-assisted product collection and MEC service provisioning, followed by the problem formulation and analysis in \cref{sec:Problem_Formulation}. Next, \cref{sec:Proposed_Framework} introduces the proposed agentic AI framework, while \cref{sec:MDP} presents the hierarchical DRL approach for solving the formulated problem. The performance of the agentic AI framework and the hierarchical DRL approach is evaluated in \cref{sec:Results}. Finally, \cref{sec:Conclusion} concludes this paper.

\section{Related Work}
\label{sec:Related_Work}
This section comprehensively reviews prior studies in three related areas, namely the UAVs routing problem, computational task offloading problem, and agentic AI. It discusses the progress achieved in these domains while outlining the limitations that motivate and position the contributions of this work. 

\subsection{Vehicle Routing Problem}
UAV routing has attracted growing attention in a range of applications, including aerial survey, logistics delivery, and maritime services. For example, the authors in \cite{Ref_Related_UAV1} studied a glider routing problem for aerial survey and jointly considered routing and flight trajectory optimization through a matheuristic framework. The authors in \cite{Ref_Related_UAV2} formulated a dynamic drone routing problem with uncertain demand and energy consumption as an MDP and developed an approximate dynamic solution for routing, drone usage, and battery management. From a learning-based perspective, the authors in \cite{Ref_Related_UAV3} proposed a multi-agent DRL framework for dynamic logistics UAV routing in large-scale multi-UAV systems. In maritime settings, a collaborative vessel-UAV routing problem with time windows for offshore delivery was discussed in \cite{Ref_Related_UAV4}. Moreover, a shore-to-ship drone routing problem with non-stationary vessel locations and non-linear energy consumption was presented in \cite{Ref_Related_UAV5}. Recent energy-aware vehicle-routing studies have also introduced learning-based methods that combine realistic energy modeling with route construction. For example, \cite{R2_Q2_Ref1} formulated a heterogeneous fleet-based capacitated electric vehicle routing problem and developed a dynamic-aware DRL method with an encoder-updater-decoder architecture to capture vehicle heterogeneity, energy consumption, and evolving routing states. In addition, \cite{R2_Q2_Ref2} studied an energy-optimal electric vehicle routing problem with pickup-delivery and time-window constraints and proposed a heterogeneous attention-driven DRL framework with a dynamic-aware context embedding for energy-aware route construction. Despite these advances in UAV routing and energy-aware vehicle routing, existing studies mainly focus on mobility planning, vehicle assignment, collection or delivery feasibility, and energy-efficient route construction. They do not consider routing formulations in which UAV visits also define time-varying MEC service windows for downstream computational tasks. This gap motivates a routing formulation that explicitly connects station visits, service-window generation, and mission feasibility in a coupled logistics-computation system. 

\subsection{Computational Task Offloading Problem}
Computational task offloading has become a key topic in edge intelligence because efficient offloading is essential for balancing latency, energy consumption, and resource utilization in dynamic computing systems. Existing studies have explored this problem from several perspectives. For example, the authors in \cite{Ref_Related_Off1} studied mobility-aware dependent task offloading in collaborative edge computing and proposed a digital twin-assisted DRL method for dynamic offloading and bandwidth-related decision making. The authors in \cite{Ref_Related_Off2} investigated delay-sensitive task offloading with edge caching and developed a soft actor-critic based scheme to jointly optimize offloading, caching, and computation/resource allocation under delay reliability constraints. Recent works have also considered collaborative and hierarchical settings. The authors in \cite{Ref_Related_Off3} proposed a digital twin-assisted multi-agent reinforcement learning framework for reliability-aware dependent task offloading in decentralized edge computing, jointly considering bandwidth allocation. In \cite{Ref_Related_Off4}, a hierarchical aerial computing framework that combined UAVs and a high-altitude platform was used to study the joint offloading, user association, and resource allocation. The authors in \cite{Ref_Related_Off5} explored the joint task offloading and resource allocation in ultra-dense MEC using a mean-field-based DRL method to make the system more scalable. 
Recent MEC studies further show that task offloading also depends on reliable edge access and realistic wireless-system modeling; secure authentication has been studied to protect IoT-enabled MEC task outsourcing \cite{Ref_Elsevier_IoT1}, while cognitive IRS-assisted UAV-MEC has been formulated with binary offloading, UAV placement, intelligent reflecting surfaces configuration, and resource allocation under hardware and spectrum-sharing constraints \cite{Ref_Elsevier_IoT2}. 
Satellite-enabled MEC studies further extend the offloading problem to remote IoT scenarios, where digital-twin semantic offloading is used to handle intermittent low-Earth orbit connectivity and limited bandwidth \cite{Ref_Elsevier_IoT3}, and reliability-aware cooperative offloading is designed to address stochastic workloads, inter-satellite task migration, and distributed resource coordination \cite{Ref_Elsevier_IoT4}. Although these studies have advanced task offloading through complex architectures and adaptive learning-based methods, they generally model service availability from the communication or edge-computing perspective. They do not directly couple task execution with manufacturing-oriented UAV product-collection routes, routing-induced service windows, and UAV energy budgets. This gap motivates an MEC scheduling formulation that is explicitly linked to the UAV-routing layer in a cloud-edge-device manufacturing system.

\subsection{Agentic AI}

Recent advances in agentic AI have extended LLM-based systems from question answering to interactive agents with reasoning, planning, retrieval, and tool-use capabilities \cite{HanwenRAG1}. In manufacturing, prior studies have explored this trend from several perspectives. For example, the authors in \cite{Ref_Related_Agent1} reviewed the concepts, capability boundaries, and challenges of agentic AI for future manufacturing. The authors in \cite{Ref_Related_Agent2} further proposed a manufacturing-oriented agentic AI framework with coordinated LLM-based agents, a unified data-model-knowledge lake, and human-in-the-loop oversight. Another line of research focuses on domain-grounded interaction through retrieval augmentation. In \cite{Ref_Related_Agent3}, the authors developed a RAG-based LLM for industrial knowledge management over heterogeneous enterprise documents. The authors in \cite{Ref_Related_Agent4} proposed an agentic AI-assisted advanced planning and scheduling (A4PS) framework that uses multi-agent LLMs, RAG, and CoT to support advanced planning and scheduling (APS) model, algorithm-code, and schedule updates under structural changes. It follows an APS standard operating procedure (SOP)-oriented workflow and retrieves APS modeling and algorithm knowledge for model/code generation. In contrast, our framework targets formulation generation from users’ natural-language descriptions for a coupled UAV-logistics and MEC scheduling problem. This task is challenging because user inputs can be unclear or incomplete, and some requirements may be implicit. Therefore, the framework must identify the underlying variables, objectives, constraints, and cross-layer couplings before the formulated problem is solved through hierarchical PPO. Specifically, our framework uses a responder-verifier workflow rather than an APS multi-agent workflow. Its RAG retrieves formulation evidence on UAV routing, task offloading, service windows, deadlines, energy budgets, and cross-layer coupling. Its CoT decomposes formulation into objective construction, UAV-routing constraints, task-offloading constraints, and notation summarization with step-level verification. In \cite{Ref_Related_Agent5}, the authors studied agentic data analysis in manufacturing through a lightweight workflow for natural-language interaction with manufacturing datasets. Although these studies highlight the value of agentic AI in manufacturing, they mainly focus on conceptual frameworks, industrial question answering, scheduling assistance, or data interaction. Its use for customized mathematical system modeling remains limited, especially for selectively retrieving domain knowledge and organizing it into an interpretable optimization formulation for tightly coupled hybrid problems. This gap motivates the framework proposed in this work.

\section{System Model}
\label{sec:SystemModel}
In this section, we provide a system overview, a UAV routing model, and a MEC service model. The relevant symbols are summarized in \cref{tab:symbols_decision_variables,tab:symbols_parameters,tab:symbols_superscripts_others}.

\begin{table}[pos=htbp]
    \centering
    \caption{Symbols: Decision Variables}
    \label{tab:symbols_decision_variables}
    \begingroup
    \setlength{\tabcolsep}{4pt}
    \renewcommand{\arraystretch}{1.08}
    \renewcommand{\tabularxcolumn}[1]{p{#1}}
    \begin{tabularx}{\textwidth}{
        @{}
        >{\raggedright\arraybackslash}p{1.2cm}
        >{\raggedright\arraybackslash}X
        >{\raggedright\arraybackslash}p{1.2cm}
        >{\raggedright\arraybackslash}X
        @{}
    }
        \toprule
        \textbf{Symbols} 
        & \textbf{Description} 
        & \textbf{Symbols} 
        & \textbf{Description} \\
        \midrule


        $b_{k,\tau,t,u}^{\mathrm{bh}}$ 	
        & UAV-$u\!\to$cloud rate allocated (bits/s). 
        & $p_{k,\tau,u}$ 	
        & Binary: task $(k,\tau)$ processed on UAV $u$. \\

        $b_{k,\tau,t,u}^{\mathrm{ul}}$ 	
        & ISD$\!\to$UAV-$u$ rate allocated (bits/s). 
        & $\zeta_{k,\tau}$ 	
        & Binary: task $(k,\tau)$ processed locally. \\

        $c_m$ 	
        & Binary: station $m$ collected. 
        & $\iota_{k,\tau,u}^{\mathrm{fh}}$  	
        & Binary: UAV $u$ is used as the first hop for task $(k,\tau)$. \\

        $\delta_{k,\tau,t}$ 	
        & Binary: task $(k,\tau)$ completes in slot $t$. 
        & $s_u,\ r_u$ 	
        & Depot departure / return time of UAV $u$. \\

        $\eta_{u,m,t}$ 	
        & Binary: UAV $u$ serves station $m$ in slot $t$. 
        & $T_{k,\tau}$ 	
        & Task completion time. \\

        $f_{k,\tau,t,u}^{\mathrm{cld}}$ 	
        & Cloud compute via UAV $u$ allocated (work-unit/s). 
        & $T_{u,m}^{\mathrm{arr}},\ T_{u,m}^{\mathrm{dep}}$ 	
        & Arrival / departure time of UAV $u$ at station $m$. \\

        $f_{k,\tau,t,u}^{\mathrm{uav}}$ 	
        & UAV-$u$ compute allocated (work-unit/s). 
        & $x_{u,i,j}$ 	
        & Binary: UAV $u$ travels $i\!\rightarrow\! j$. \\

        $f_{k,\tau,t}^{\mathrm{loc}}$ 	
        & Local compute allocated (work-unit/s). 
        & $y_{u,m}$ 	
        & Binary: station $m$ assigned to UAV $u$. \\

        $g_{k,\tau,u}$ 	
        & Binary: task $(k,\tau)$ executed in cloud via UAV $u$. 
        & $z_{k,\tau}$ 	
        & Binary: task $(k,\tau)$ meets deadline. \\

        \bottomrule
    \end{tabularx}
    \endgroup
\end{table}

\begin{table}[pos=htbp]
    \centering
    \caption{Symbols: Parameters}
    \label{tab:symbols_parameters}
    \begingroup
    \setlength{\tabcolsep}{4pt}
    \renewcommand{\arraystretch}{1.08}
    \renewcommand{\tabularxcolumn}[1]{p{#1}}
    \begin{tabularx}{\textwidth}{
        @{}
        >{\raggedright\arraybackslash}p{1.2cm}
        >{\raggedright\arraybackslash}X
        >{\raggedright\arraybackslash}p{1.4cm}
        >{\raggedright\arraybackslash}X
        @{}
    }
        \toprule
        \textbf{Symbols} 
        & \textbf{Description} 
        & \textbf{Symbols} 
        & \textbf{Description} \\
        \midrule


        $\alpha_{\mathrm{bh}}$	
        & Energy coefficient for UAV-to-cloud communication. 
        & $F^{\mathrm{cld}}$ 	
        & Cloud compute capacity (work-unit/s). \\

        $\alpha_{\mathrm{cmp}}$	
        & Energy coefficient for onboard computing. 
        & $F_k^{\mathrm{loc}}$ 	
        & ISD $k$ local compute capacity (work-unit/s). \\

        $\alpha_{\mathrm{fly}},\ \alpha_{\mathrm{hov}}$	
        & Energy coefficients for flying and hovering. 
        & $F_u^{\mathrm{uav}}$ 	
        & UAV $u$ compute capacity (work-unit/s). \\

        $\alpha_{\mathrm{ul}}$	
        & Energy coefficient for ISD-to-UAV communication. 
        & $\omega_{\mathrm{col}},\ \omega_{\mathrm{cmp}}$	
        & Weights for collection value and task completion. \\

        $B_{k,\tau}$ 	
        & Task input size (bits). 
        & $\omega_{\mathrm{miss}},\ \omega_{\mathrm{flow}}$	
        & Weights for miss penalty and flow-time. \\

        $B_u^{\mathrm{ul}},\ B_u^{\mathrm{bh}}$ 	
        & UAV $u$ uplink / backhaul limits (bits/s). 
        & $\omega_{\mathrm{res}}$	
        & Weight for resource-occupation cost. \\

        $\gamma_{k,t,u}^{\mathrm{ul}},\ \gamma_{t,u}^{\mathrm{bh}}$ 	
        & Effective-rate factors (uplink / backhaul). 
        & $Q_u$ 	
        & UAV payload capacity. \\

        $\Delta$ 	
        & Time slot duration. 
        & $\tau_m$ 	
        & Minimum service time at station $m$. \\

        $d_{i,j}$ 	
        & Distance between nodes $i$ and $j$. 
        & $T_{\mathrm{mission}}$ 	
        & Mission horizon. \\

        $D_k$ 	
        & Task deadline duration for ISD $k$ tasks. 
        & $v_{\mathrm{fly}}$ 	
        & UAV flight speed. \\

        $D_u$ 	
        & UAV distance budget. 
        & $v_m,\ w_m$ 	
        & Product value / weight at station $m$. \\

        $E_u^{\max}$ 	
        & UAV $u$ energy budget. 
        & $W_{k,\tau}$ 	
        & Task workload (work-units). \\

        \bottomrule
    \end{tabularx}
    \endgroup
\end{table}

\begin{table}[pos=htbp]
    \centering
    \caption{Symbols: Superscripts and Notations}
    \label{tab:symbols_superscripts_others}
    \begingroup
    \setlength{\tabcolsep}{4pt}
    \renewcommand{\arraystretch}{1.08}
    \renewcommand{\tabularxcolumn}[1]{p{#1}}
    \begin{tabularx}{\textwidth}{
        @{}
        >{\raggedright\arraybackslash}p{1.55cm}
        >{\raggedright\arraybackslash}X
        >{\raggedright\arraybackslash}p{1.35cm}
        >{\raggedright\arraybackslash}X
        @{}
    }
        \toprule
        \textbf{Symbols} 
        & \textbf{Description} 
        & \textbf{Symbols} 
        & \textbf{Description} \\
        \midrule

        $\mathrm{cmp},\ \mathrm{com}$ 	
        & Computing-related / communication-related. 
        & $L$ 	
        & Set of routing nodes, $L=\{0\}\cup S$, where $0$ denotes the depot. \\

        $\mathrm{fly},\ \mathrm{hov}$ 	
        & Flight / hovering energy tag. 
        & $m(k)$ 	
        & Station hosting ISD $k$. \\

        $\mathrm{loc},\ \mathrm{uav},\ \mathrm{cld}$ 	
        & Local / UAV / cloud mode or capacity type. 
        & $m,u,k,i,j$ 	
        & Indices: station, UAV, ISD, node $i$, node $j$. \\

        $\mathrm{ul},\ \mathrm{bh}$ 	
        & Uplink (ISD$\!\to$UAV) / backhaul (UAV$\!\to$cloud). 
        & $N_{\mathrm{slot}}$ 	
        & Total number of time slots in $\mathcal{T}=\{0,\dots,N_{\mathrm{slot}}-1\}$. \\

        $\max$ 	
        & Budget upper bound. 
        & $R_t^{\mathrm{cmp}}$ 	
        & Normalized computing occupation in slot $t$. \\

        $\chi_{k,\tau}(t)$ 	
        & Cumulative completion flag by slot $t$. 
        & $R_t^{\mathrm{com}}$ 	
        & Normalized communication occupation in slot $t$. \\

        $E_u^{\mathrm{cmp}},\ E_u^{\mathrm{com}}$ 	
        & UAV onboard-computing / communication energy. 
        & $S,\ U,\ K$ 	
        & Sets of stations / UAVs / ISDs. \\

        $E_u^{\mathrm{fly}}$ 	
        & UAV flight-hover energy. 
        & $\mathcal{T}$ 	
        & Time slot set; $t\in\mathcal{T}$. \\

        $\mathcal{I}$ 	
        & Realized-task set indexed by $(k,\tau)$. 
        & $\tau,t$ 	
        & Indices: task-generation slot / slot. \\

        \bottomrule
    \end{tabularx}
    \endgroup
\end{table}

\begin{figure}[pos=htbp]    
  \centering
  \includegraphics[width=0.5\columnwidth]{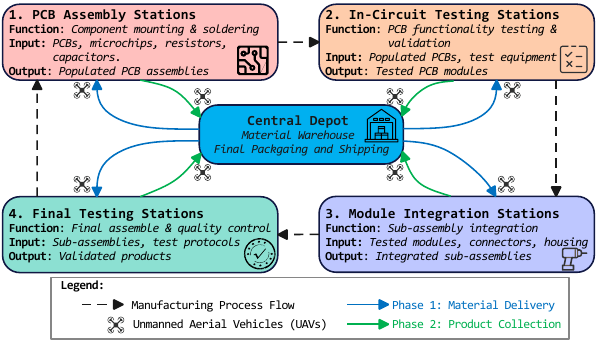}
    \caption{An example of UAV-assisted electronics manufacturing with a two-phase logistics coordination system. The manufacturing process flow includes PCB assembly, in-circuit testing, module integration, and final testing. The UAVs handle logistics in two phases: delivering materials before manufacturing in phase 1 and collecting products after manufacturing in phase 2.}
  \label{fig:electro_Mag}
\end{figure}

\subsection{System Overview}
A typical manufacturing sequence consists of material delivery, manufacturing completion, and product collection. Accordingly, the logistics process has two phases, namely phase 1 for material delivery and phase 2 for product collection. In this paper, we focus on phase 2. Specifically, a fleet of $U$ homogeneous UAVs is dispatched from a central depot equipped with terrestrial MEC servers, hereafter referred to as the cloud, to collect finished products from $M$ geographically distributed manufacturing stations. At the same time, these UAVs provide MEC services to industrial sensor devices (ISDs) deployed at the stations. All UAVs depart from and return to the depot, and they fly at a fixed altitude with constant speed. Each station $m \in S$ is located at $(X_m,Y_m)$, has a collection reward $v_m$, product weight $w_m$, and hosts a set of ISDs $K_m$. All finished products are assumed to be ready for collection at the mission start, while ISDs generate computational tasks stochastically over a discretized time horizon. 

The considered system couples physical logistics with computational task processing. 
On the logistics side\footnote{\url{https://wing.com/}}, each UAV is constrained by payload, travel distance, mission time, and battery energy, and must complete a depot-based collection mission \cite{Ref_joint_LC1}. On the computing side, task execution follows a three-layer cloud-edge-device architecture, where a task can be processed locally at the ISD, at a serving UAV, or at the cloud via a UAV-assisted path \cite{sun_UVA}. Since UAV-assisted processing is available only when a UAV is serving the corresponding station, routing decisions directly determine the service windows for MEC tasks \cite{Ref_hyb2}. Therefore, the system is modeled from two coupled perspectives in the following subsections: the \emph{UAV Routing Model}, which captures UAV collection decisions and routing feasibility, and the \emph{MEC Service Model}, which captures task execution and resource-allocation decisions under the routing-induced service availability. 

A representative application of the proposed system model is the electronics component assembly and testing industry, where geographically distributed stations perform operations such as PCB assembly, in-circuit testing, module integration, and final testing. As illustrated in \Cref{fig:electro_Mag}, UAVs support two-phase logistics: They first deliver raw materials or sub-assembled components to the stations and then collect finished products for transporting to a central depot after processing is completed \cite{UseCase2_R3}. During both phases, the UAVs can also provide MEC support by executing offloaded tasks on board or relaying them to the cloud, thereby supporting timely processing of industrial sensing and analytics tasks. Such UAV-assisted operation is well suited to modern manufacturing environments, where UAVs are increasingly used for delivery, monitoring, inspection, inventory management, and predictive maintenance \cite{UseCase2_R13b}. In indoor manufacturing environments, UAVs offer spatial flexibility in occupied or hard-to-access areas and can operate in three-dimensional workspaces, making them suitable for spatially constrained and complex industrial settings \cite{UseCase2_R12_update,UseCase2_R13}. 

\subsection{UAV Routing Model} 
The UAV-routing constraints are organized into three parts, namely collection assignment and route consistency, payload and distance budgets, and mission timing feasibility. 

\subsubsection{Collection assignment and route consistency}

\begin{align}
\label{eq:C1_aligned}
&\sum_{u\in U} y_{u,m}\le 1, \quad
c_m\le \sum_{u\in U} y_{u,m},
&& \forall m\in S,\\
\label{eq:C3_aligned}
&\sum_{i\in L} x_{u,i,m}
=
\sum_{j\in L} x_{u,m,j}
=
y_{u,m},
&&\forall u\in U,\ \forall m\in S,\\
\label{eq:C4_aligned}
&\sum_{m\in S} x_{u,0,m}
=
\sum_{m\in S} x_{u,m,0},
&&\forall u\in U, \\
\label{eq:C5_no_self_loop}
& x_{u,i,i}=0, 
&& \forall u\in U,\ \forall i\in L. 
\end{align}
\Cref{eq:C1_aligned} limits each station to at most one UAV and allows a station to be counted as collected only if it is assigned. 
\Cref{eq:C3_aligned} ensures that station $m$ appears on the route of UAV $u$ exactly when it is assigned to UAV $u$, in which case it is entered and left exactly once by UAV $u$. 
\Cref{eq:C4_aligned} ensures that each UAV departs from and returns to the depot the same number of times. 
\Cref{eq:C5_no_self_loop} excludes self-loops at both the depot and stations. 
The constraints in \cref{eq:C1_aligned,eq:C3_aligned,eq:C4_aligned} define the assignment, route-consistency, and depot-balance conditions, while disconnected station subtours are eliminated by the mission-timing constraints, as discussed in \cref{subsubsec_subtour_elimination}. 

\subsubsection{Payload and distance budgets}

\begin{equation}
\label{eq:C6C7_aligned}
\sum_{m\in S} w_m y_{u,m}\le Q_u,   \,\,\,\,
\sum_{i\in L}\sum_{j\in L} d_{i,j}x_{u,i,j}\le D_u, \,\,\,
\forall u\in U.
\end{equation}
\Cref{eq:C6C7_aligned} limits the total assigned payload weight and the total travel distance of each UAV.

\subsubsection{Mission timing}
\label{subsubsec_Mission_timing}

Let $M_T$ be a sufficiently large time-domain constant satisfying $M_T \ge T_{\mathrm{mission}}+\max_{i,j\in L} d_{i,j}/v_{\mathrm{fly}}$. The mission-timing constraints are formulated using the following linear Big-$M$ constraints: 
\begin{align}
\label{eq:depot_dep_horizon_merged}
    &s_u=0,\qquad r_u\le T_{\mathrm{mission}},
    &&\forall u\in U,\\
    \label{eq:travel_time_merged1}
    & T_{u,m}^{\mathrm{arr}}
    \ge 
    s_u+\frac{d_{0,m}}{v_{\mathrm{fly}}}
    -M_T(1-x_{u,0,m}),    
    &&\forall u\in U,\ \forall m\in S,\\
    \label{eq:travel_time_merged2}	   
    & T_{u,j}^{\mathrm{arr}}
    \ge 
    T_{u,i}^{\mathrm{dep}}+\frac{d_{i,j}}{v_{\mathrm{fly}}}
     -M_T(1-x_{u,i,j}),    
    &&   \forall u\in U,\ \forall i,j\in S,\ i\neq j,   
    \\
    \label{eq:travel_time_merged3}
    &  r_u
    \ge 
    T_{u,m}^{\mathrm{dep}}+\frac{d_{m,0}}{v_{\mathrm{fly}}}
     -M_T(1-x_{u,m,0}),   
    &&\forall u\in U,\ \forall m\in S,\\
    \label{eq:C17_aligned}
    &T_{u,m}^{\mathrm{dep}}-T_{u,m}^{\mathrm{arr}}\ge \tau_m\, y_{u,m},
    &&\forall u\in U,\ \forall m\in S.
\end{align}
\Cref{eq:depot_dep_horizon_merged} states that each UAV starts at time $0$ and must return within the mission horizon. \Cref{eq:travel_time_merged1,eq:travel_time_merged2,eq:travel_time_merged3} are linearized formulations of the route-timing implications. They enforce the corresponding travel-time relation when the associated arc is selected, and relax it otherwise through the Big-$M$ term. In \Cref{eq:travel_time_merged2}, the case $i=j$ is excluded, since no travel is needed from a station to itself. \Cref{eq:C17_aligned} requires a minimum service duration at each station assigned to UAV $u$. This service interval later determines the feasible slots in which UAV-assisted MEC service can be made available.

\subsubsection{Discussion on the subtour-elimination property}
\label{subsubsec_subtour_elimination}

The timing constraints in \cref{subsubsec_Mission_timing} also eliminate disconnected station subtours. Suppose that a disconnected station subtour $m_1\rightarrow m_2\rightarrow \cdots \rightarrow m_q\rightarrow m_1$ exists for UAV $u$, where $q\ge 2$, $m_1,\dots,m_q\in S$ are distinct stations, and $m_{q+1}=m_1$. For every selected arc in this subtour, i.e., $x_{u,m_\ell,m_{\ell+1}}=1$, \cref{eq:travel_time_merged2} imposes 
\[
T_{u,m_{\ell+1}}^{\mathrm{arr}}
\ge
T_{u,m_{\ell}}^{\mathrm{dep}}
+
\frac{d_{m_\ell,m_{\ell+1}}}{v_{\mathrm{fly}}},
\qquad \ell=1,\dots,q.
\]
Because each station in the subtour is visited by UAV $u$, 
\cref{eq:C3_aligned} gives $y_{u,m_\ell}=1$ for all $\ell=1,\dots,q$. 
Together with the service-duration constraint in \cref{eq:C17_aligned}, this gives
\[
T_{u,m_{\ell+1}}^{\mathrm{arr}}
\ge
T_{u,m_{\ell}}^{\mathrm{arr}}
+
\tau_{m_\ell}
+
\frac{d_{m_\ell,m_{\ell+1}}}{v_{\mathrm{fly}}},
\qquad \ell=1,\dots,q.
\]
Summing these inequalities over the closed cycle yields
\[
0
\ge
\sum_{\ell=1}^{q}
\left(
\tau_{m_\ell}
+
\frac{d_{m_\ell,m_{\ell+1}}}{v_{\mathrm{fly}}}
\right).
\]
This is impossible because the right-hand side is strictly positive for distinct stations with positive travel times and nonnegative service durations. 
Therefore, disconnected station subtours are infeasible under the mission-timing constraints.

\subsection{MEC Service Model}  
The constraints associated with MEC service model are about task offloading, including execution-mode selection, completion-time encoding, computing and communication resource limits, service-window indicators, service-window feasibility for UAV-involved processing, deadline satisfaction, and UAV energy budgeting. 

\subsubsection{Execution-mode selection}
\begin{equation}
\label{eq:mode_select_aligned}
\zeta_{k,\tau}
+\sum_{u\in U} p_{k,\tau,u}
+\sum_{u\in U} g_{k,\tau,u}
=1,
\qquad
\forall (k,\tau)\in\mathcal{I}.
\end{equation}
\Cref{eq:mode_select_aligned} requires every task to choose exactly one execution mode: local execution, UAV execution, or cloud execution via a UAV relay.

\subsubsection{Completion-time encoding}
\begin{align}
\label{eq:Tcomp_def1}
z_{k,\tau}
&=
\sum_{t=\tau}^{N_{\mathrm{slot}}-1}\delta_{k,\tau,t},
\qquad
\forall (k,\tau)\in\mathcal{I}, \\
\label{eq:Tcomp_def2}
T_{k,\tau}
&=
\sum_{t=\tau}^{N_{\mathrm{slot}}-1}(t+1)\Delta\,\delta_{k,\tau,t}
+T_{\mathrm{mission}}(1-z_{k,\tau}). 
\end{align}
\Cref{eq:Tcomp_def1,eq:Tcomp_def2} define the completion-status indicator $z_{k,\tau}$ and the completion time $T_{k,\tau}$ based on the slot-selection binary $\delta_{k,\tau,t}\in\{0,1\}$. Specifically, $\delta_{k,\tau,t}=1$ indicates that task $(k,\tau)$ is completed by the end of slot $t$. 

\subsubsection{Computing and communication resource constraints}

The computing and communication resources are subject to three types of constraints, namely per-slot capacity limits, mode consistency, and cumulative service sufficiency.

\textbf{(i) Per-slot capacity limits}
\begin{align}
\label{eq:cap_compute}
&\left\{
\begin{aligned}
&\sum_{\tau':(k,\tau')\in\mathcal{I},\ \tau'\le t}
f_{k,\tau',t}^{\mathrm{loc}}
\le F_k^{\mathrm{loc}},
&&\forall k\in K,\ \forall t\in\mathcal{T},\\
&\sum_{(k,\tau)\in\mathcal{I}:\tau\le t}
f_{k,\tau,t,u}^{\mathrm{uav}}
\le F_u^{\mathrm{uav}},
&&\forall u\in U,\ \forall t\in\mathcal{T},\\
&\sum_{u\in U}\sum_{(k,\tau)\in\mathcal{I}:\tau\le t}
f_{k,\tau,t,u}^{\mathrm{cld}}
\le F^{\mathrm{cld}},
&&\forall t\in\mathcal{T}.
\end{aligned}
\right.
\\[2pt]
\label{eq:cap_comm}
&\left\{
\begin{aligned}
&\sum_{(k,\tau)\in\mathcal{I}:\tau\le t}
b_{k,\tau,t,u}^{\mathrm{ul}}
\le B_u^{\mathrm{ul}},
&&\forall u\in U,\ \forall t\in\mathcal{T},\\
&\sum_{(k,\tau)\in\mathcal{I}:\tau\le t}
b_{k,\tau,t,u}^{\mathrm{bh}}
\le B_u^{\mathrm{bh}},
&&\forall u\in U,\ \forall t\in\mathcal{T}.
\end{aligned}
\right.
\end{align}
\Cref{eq:cap_compute,eq:cap_comm} bound the aggregate computing and communication resource allocations in each slot by the available local, UAV, cloud, uplink, and backhaul capacities.

\textbf{(ii) Mode Consistency}
\begin{align}
\label{eq:mode_gate_comp}
& \left\{
\begin{aligned}
&0\le f_{k,\tau,t}^{\mathrm{loc}}
\le F_k^{\mathrm{loc}}\,\zeta_{k,\tau},
&&\forall (k,\tau)\in\mathcal{I},\ \forall t\ge \tau,\\
&0\le f_{k,\tau,t,u}^{\mathrm{uav}}
\le F_u^{\mathrm{uav}}\,p_{k,\tau,u},
&&\forall (k,\tau)\in\mathcal{I},\ \forall u\in U,\ \forall t\ge \tau,\\
&0\le f_{k,\tau,t,u}^{\mathrm{cld}}
\le F^{\mathrm{cld}}\,g_{k,\tau,u},
&&\forall (k,\tau)\in\mathcal{I},\ \forall u\in U,\ \forall t\ge \tau,
\end{aligned} 
\right. 
\\[2pt] 
\label{eq:mode_gate_comm}
& \left\{
\begin{aligned}
&0\le b_{k,\tau,t,u}^{\mathrm{ul}}
\le B_u^{\mathrm{ul}}\,\iota_{k,\tau,u}^{\mathrm{fh}}, \quad\,\,\,\,
\forall (k,\tau)\in\mathcal{I},\ \forall u\in U,\ \forall t\ge \tau,\\
&0\le b_{k,\tau,t,u}^{\mathrm{bh}}
\le B_u^{\mathrm{bh}}\,g_{k,\tau,u}, \quad\,\, 
\forall (k,\tau)\in\mathcal{I},\ \forall u\in U,\ \forall t\ge \tau,\\
&\iota_{k,\tau,u}^{\mathrm{fh}}=p_{k,\tau,u}+g_{k,\tau,u}.
\end{aligned}
\right.
\end{align} 
\Cref{eq:mode_gate_comp,eq:mode_gate_comm} impose mode consistency on computing and communication allocations by requiring each allocation to be zero unless the corresponding execution mode is selected.

\textbf{(iii) Multi-slot sufficiency for completion}
Define
$\chi_{k,\tau}(t)\triangleq \sum_{r=\tau}^{t}\delta_{k,\tau,r}\in\{0,1\}$, 
which indicates whether task $(k,\tau)$ has been completed by the end of slot $t$. 
Let $M_W$ and $M_B$ be sufficiently large constants satisfying 
$M_W\ge \max_{(k,\tau)\in\mathcal{I}} W_{k,\tau}$ and 
$M_B\ge \max_{(k,\tau)\in\mathcal{I}} B_{k,\tau}$. 
For any $t\ge \tau$, the cumulative service must be sufficient whenever completion by slot $t$ is claimed. 
The corresponding Big-$M$ linear constraints are:
\begin{align}
\label{eqn:suff}
&\left\{
\begin{aligned}
&\sum_{s=\tau}^{t} f_{k,\tau,s}^{\mathrm{loc}}\Delta
\ge 
W_{k,\tau}\chi_{k,\tau}(t) -M_W(1-\zeta_{k,\tau}), 
&& \forall (k,\tau)\in\mathcal{I},\ \forall t\ge \tau,\\
&\sum_{s=\tau}^{t} f_{k,\tau,s,u}^{\mathrm{uav}}\Delta
\ge 
W_{k,\tau}\chi_{k,\tau}(t) -M_W(1-p_{k,\tau,u}),  
&& \forall (k,\tau)\in\mathcal{I},\ \forall u\in U,\ \forall t\ge \tau,\\
&\sum_{s=\tau}^{t} f_{k,\tau,s,u}^{\mathrm{cld}}\Delta
\ge 
W_{k,\tau}\chi_{k,\tau}(t)  -M_W(1-g_{k,\tau,u}),  
&& \forall (k,\tau)\in\mathcal{I},\ \forall u\in U,\ \forall t\ge \tau.
\end{aligned}
\right.
\\[2pt]
\label{eqn:suff4}
& \left\{
\begin{aligned}
&\sum_{s=\tau}^{t}
\gamma_{k,s,u}^{\mathrm{ul}}\, b_{k,\tau,s,u}^{\mathrm{ul}}\,\Delta
\ge 
B_{k,\tau}\chi_{k,\tau}(t) -M_B(1-\iota_{k,\tau,u}^{\mathrm{fh}}), 
&& \forall (k,\tau)\in\mathcal{I},\ \forall u\in U,\ \forall t\ge \tau,\\
&\sum_{s=\tau}^{t}
\gamma_{s,u}^{\mathrm{bh}}\, b_{k,\tau,s,u}^{\mathrm{bh}}\,\Delta
\ge 
B_{k,\tau}\chi_{k,\tau}(t) -M_B(1-g_{k,\tau,u}), 
&& \forall (k,\tau)\in\mathcal{I},\ \forall u\in U,\ \forall t\ge \tau.								
\end{aligned}
\right. 
\end{align}
\Cref{eqn:suff} ensures that a task can be declared completed only after receiving enough cumulative computing service under its selected execution mode. 
\Cref{eqn:suff4} imposes a corresponding sufficiency constraint on communication: the cumulative transmitted bits must be no smaller than the task input size whenever completion is claimed and the corresponding UAV-assisted communication path is selected. 
The Big-$M$ terms relax the corresponding sufficiency constraints when the associated execution or communication mode is not selected.

\subsubsection{Service-window indicators}
\begingroup 
\setlength{\abovedisplayskip}{3pt}
\setlength{\belowdisplayskip}{3pt}
\setlength{\abovedisplayshortskip}{3pt}
\setlength{\belowdisplayshortskip}{3pt}
\begin{align}
\label{eq:eta_one_station_aligned}
&\sum_{m\in S}\eta_{u,m,t}\le 1,
&&\forall u\in U,\ \forall t\in\mathcal{T},\\
\label{eq:eta_assign_aligned}
&\eta_{u,m,t}\le y_{u,m},
&&\forall u\in U,\ \forall m\in S,\ \forall t\in\mathcal{T},\\
\label{eq:eta_time_merged_indicator_1}
& T_{u,m}^{\mathrm{arr}}
\le 
t\Delta + M_T(1-\eta_{u,m,t}),  
&&\forall u\in U,\ \forall m\in S,\ \forall t\in\mathcal{T},\\
\label{eq:eta_time_merged_indicator_2}
&  (t+1)\Delta
\le 
T_{u,m}^{\mathrm{dep}} + M_T(1-\eta_{u,m,t}), 
&&\forall u\in U,\ \forall m\in S,\ \forall t\in\mathcal{T}.
\end{align}
\endgroup
\Cref{eq:eta_one_station_aligned,eq:eta_assign_aligned,eq:eta_time_merged_indicator_1,eq:eta_time_merged_indicator_2} define the slot-level service-window indicators by allowing service only at assigned stations and only during the corresponding arrival-departure interval. 
Specifically, $\eta_{u,m,t}=1$ means that UAV $u$ is available to provide MEC service to station $m$ during the whole slot $[t\Delta,(t+1)\Delta]$. 
Therefore, the route-induced service duration $T_{u,m}^{\mathrm{dep}}-T_{u,m}^{\mathrm{arr}}$ determines the feasible slots in which UAV-assisted processing and cloud offloading through the UAV-assisted path can be used.

\subsubsection{Service-window feasibility for UAV-involved processing}
Let $m(k)$ denote the station to which ISD $k$ belongs. Then, 
\begin{align}
\label{eq:sw_ul_gate_aligned}
&b_{k,\tau,t,u}^{\mathrm{ul}}
\le B_u^{\mathrm{ul}}\,\eta_{u,m(k),t},
\quad
f_{k,\tau,t,u}^{\mathrm{uav}}
\le F_u^{\mathrm{uav}}\,\eta_{u,m(k),t},
\quad b_{k,\tau,t,u}^{\mathrm{bh}}
\le B_u^{\mathrm{bh}}\,\eta_{u,m(k),t},
\quad
\forall (k,\tau)\in\mathcal{I},\ \forall u\in U,\ \forall t\ge \tau.
\end{align} 
\Cref{eq:sw_ul_gate_aligned} requires the UAV-involved communication and computing allocations of task $(k,\tau)$ in slot $t$ to be zero unless UAV $u$ is serving station $m(k)$ in that slot.

\subsubsection{Deadline constraint}
\begin{equation}
\label{eq:deadline_indicator}
T_{k,\tau}
\le 
\tau\Delta + D_k + M_T(1-z_{k,\tau}), 
\qquad
\forall (k,\tau)\in\mathcal{I}.
\end{equation}
\Cref{eq:deadline_indicator} is the Big-$M$ linearized formulation of the deadline constraint. 
If $z_{k,\tau}=1$, task $(k,\tau)$ must be completed no later than its deadline $\tau\Delta+D_k$. 
If $z_{k,\tau}=0$, the constraint is relaxed by the Big-$M$ term.

\subsubsection{UAV energy budget}
The total energy of UAV $u$ is decomposed into flight/hover energy, onboard computing energy, and communication energy:

\begin{align}
\label{eqn_energy1}
E_u^{\mathrm{fly}}
&=
\alpha_{\mathrm{fly}}
\sum_{i\in L}\sum_{j\in L} d_{i,j}x_{u,i,j} +
\alpha_{\mathrm{hov}}
\sum_{m\in S}(T_{u,m}^{\mathrm{dep}}-T_{u,m}^{\mathrm{arr}})y_{u,m}, \\
\label{eqn_energy2}
E_u^{\mathrm{cmp}}
&=
\alpha_{\mathrm{cmp}}
\sum_{t\in\mathcal{T}}
\sum_{(k,\tau)\in\mathcal{I}:\tau\le t}
f_{k,\tau,t,u}^{\mathrm{uav}}\Delta,\\
\label{eqn_energy3}
E_u^{\mathrm{com}}
&=
\alpha_{\mathrm{ul}}
\sum_{t\in\mathcal{T}}
\sum_{(k,\tau)\in\mathcal{I}:\tau\le t}
\gamma_{k,t,u}^{\mathrm{ul}}\, b_{k,\tau,t,u}^{\mathrm{ul}}\,\Delta
+
\alpha_{\mathrm{bh}}
\sum_{t\in\mathcal{T}}
\sum_{(k,\tau)\in\mathcal{I}:\tau\le t}
\gamma_{t,u}^{\mathrm{bh}}\, b_{k,\tau,t,u}^{\mathrm{bh}}\,\Delta.
\end{align}
The total energy must satisfy
\begin{align}
\label{eqn_energy4}
&E_u^{\mathrm{fly}}+E_u^{\mathrm{cmp}}+E_u^{\mathrm{com}}
\le E_u^{\max}, \qquad \forall u\in U.
\end{align}
\Cref{eqn_energy1,eqn_energy2,eqn_energy3} define the flight/hover, onboard computing, and communication energy components of UAV $u$, and \Cref{eqn_energy4} imposes the corresponding per-UAV energy budget over the whole mission. 
In \cref{eqn_energy3}, $\gamma_{k,t,u}^{\mathrm{ul}}$ and 
$\gamma_{t,u}^{\mathrm{bh}}$ are dimensionless effective-rate factors for the 
ISD-to-UAV uplink and the UAV-to-cloud backhaul, respectively. They capture 
the effect of wireless channel conditions on the effective transmitted bits. 
Specifically, we define
\begin{align}
\gamma_{k,t,u}^{\mathrm{ul}}
&=
\min\left\{1,\frac{C_{k,t,u}^{\mathrm{ul}}}{B_u^{\mathrm{ul}}}\right\},
& \text{where} \,\,\,\,
C_{k,t,u}^{\mathrm{ul}}
&=
W^{\mathrm{ul}}
\log_2\!\left(
1+
\frac{P_k^{\mathrm{ul}}G_{k,t,u}^{\mathrm{ul}}}{\sigma_{\mathrm{ul}}^2}
\right),
\\
\gamma_{t,u}^{\mathrm{bh}}
&=
\min\left\{1,\frac{C_{t,u}^{\mathrm{bh}}}{B_u^{\mathrm{bh}}}\right\},
& \text{where} \quad\,\,
C_{t,u}^{\mathrm{bh}}
&=
W^{\mathrm{bh}}
\log_2\!\left(
1+
\frac{P_u^{\mathrm{bh}}G_{t,u}^{\mathrm{bh}}}{\sigma_{\mathrm{bh}}^2}
\right).
\end{align}
Here, $C_{k,t,u}^{\mathrm{ul}}$ and $C_{t,u}^{\mathrm{bh}}$ denote the channel-achievable uplink and backhaul rates in slot $t$, respectively. $W^{\mathrm{ul}}$ and $W^{\mathrm{bh}}$ denote the corresponding channel bandwidths, respectively. $P_k^{\mathrm{ul}}$ and $P_u^{\mathrm{bh}}$ denote the transmit powers of ISD $k$ and UAV $u$, respectively. $G_{k,t,u}^{\mathrm{ul}}$ and $G_{t,u}^{\mathrm{bh}}$ denote the corresponding channel power gains, and $\sigma_{\mathrm{ul}}^2$ and $\sigma_{\mathrm{bh}}^2$ denote the noise powers. Therefore, $\gamma_{k,t,u}^{\mathrm{ul}} b_{k,\tau,t,u}^{\mathrm{ul}}\Delta$ and $\gamma_{t,u}^{\mathrm{bh}} b_{k,\tau,t,u}^{\mathrm{bh}}\Delta$ represent the effective transmitted bits over the uplink and backhaul in slot $t$, respectively. In the fixed-capacity setting used in the simulations, the channel effect can be absorbed into $B_u^{\mathrm{ul}}$ and $B_u^{\mathrm{bh}}$, which means that the configured link limits are treated as effective link capacities. In \cref{eqn_energy3}, since the $\gamma$ terms are unitless and $b\Delta$ has the unit of bits, $\alpha_{\mathrm{ul}}$ and $\alpha_{\mathrm{bh}}$ have units of $\mathrm{J/bit}$. They represent the communication energy consumed per effective transmitted bit on the ISD-to-UAV uplink and UAV-to-cloud backhaul, respectively.

\section{Problem Formulation and Analysis}
\label{sec:Problem_Formulation}
In this section, we provide a problem formulation and a problem analysis. 

\subsection{Problem Formulation}
\label{sub: problem_formulation}
We aim to maximize the overall system performance, subject to the UAV routing constraints in \cref{eq:C1_aligned}--\cref{eq:C17_aligned} and the MEC service (i.e., task offloading) constraints in \cref{eq:mode_select_aligned}--\cref{eqn_energy4}. Specifically, the objective promotes product collection and deadline-compliant task completion, while reducing deadline violations, task completion delay, and overall computing and communication resource occupation. The complete optimization problem is formulated as follows: 
\begin{align}
\label{eq:obj_revised}
\mathbf{P}: \quad
& \max \;
\underbrace{\omega_{\mathrm{col}}
\sum_{m\in S} v_m c_m}_{\text{collection value}}
+
\underbrace{\omega_{\mathrm{cmp}}
\sum_{(k,\tau)\in\mathcal{I}} z_{k,\tau}}_{\text{on-time completion return}}
-\underbrace{\omega_{\mathrm{miss}}
\sum_{(k,\tau)\in\mathcal{I}} (1-z_{k,\tau})}_{\text{deadline violation cost}}
-\underbrace{\omega_{\mathrm{flow}}
\sum_{(k,\tau)\in\mathcal{I}}
\bigl(T_{k,\tau}-\tau\Delta\bigr)}_{\text{flow-time cost}}
\notag\\
&
-\underbrace{\omega_{\mathrm{res}}
\sum_{t\in\mathcal{T}}
\bigl(R_t^{\mathrm{cmp}}+R_t^{\mathrm{com}}\bigr)}_{\text{resource-occupation cost}}
\\
\text{s.t.}\quad
& \cref{eq:C1_aligned}-\cref{eq:C17_aligned}, \qquad \cref{eq:mode_select_aligned}-\cref{eqn_energy4}. \notag
\end{align}
where $T_{k,\tau}$ denotes the completion time of task $(k,\tau)$, $z_{k,\tau}\in\{0,1\}$ indicates whether the task is completed before its deadline. 
Let $\mathcal{I}_t=\{(k,\tau)\in\mathcal{I}:\tau\le t\}$ denote the set of tasks that have arrived by slot $t$. The terms $R_t^{\mathrm{cmp}}$ and $R_t^{\mathrm{com}}$ denote the aggregated normalized computing and communication occupation in slot $t$, respectively, and are defined as
\begin{align}
\label{eqn_R_cmp}
R_t^{\mathrm{cmp}}
&=
\frac{1}{\lvert K\rvert}
\sum_{k\in K}
\frac{
\sum_{\tau:(k,\tau)\in\mathcal{I}_t}
f_{k,\tau,t}^{\mathrm{loc}}
}{
F_k^{\mathrm{loc}}
}
+
\frac{1}{\lvert U\rvert}
\sum_{u\in U}
\frac{
\sum_{(k,\tau)\in\mathcal{I}_t}
f_{k,\tau,t,u}^{\mathrm{uav}}
}{
F_u^{\mathrm{uav}}
}
+
\frac{
\sum_{u\in U}
\sum_{(k,\tau)\in\mathcal{I}_t}
f_{k,\tau,t,u}^{\mathrm{cld}}
}{
F^{\mathrm{cld}}
},
\\ 
\label{eqn_R_com}
R_t^{\mathrm{com}}
&=
\frac{1}{\lvert U\rvert}
\sum_{u\in U}
\frac{
\sum_{(k,\tau)\in\mathcal{I}_t}
b_{k,\tau,t,u}^{\mathrm{ul}}
}{
B_u^{\mathrm{ul}}
}
+
\frac{
\sum_{u\in U}
\sum_{(k,\tau)\in\mathcal{I}_t}
b_{k,\tau,t,u}^{\mathrm{bh}}
}{
\sum_{u\in U}B_u^{\mathrm{bh}}
}.
\end{align}
Thus, $R_t^{\mathrm{cmp}}$ aggregates the normalized occupation of local, UAV, and cloud computing resources, while $R_t^{\mathrm{com}}$ aggregates the normalized occupation of ISD-to-UAV uplink and UAV-to-cloud backhaul resources. These terms are unitless because each allocated resource is normalized by its corresponding capacity. Therefore, the resource-occupation term penalizes excessive per-slot usage of computing and communication resources.

\subsection{Problem Analysis}      
The mathematical formulation presented in \cref{sub: problem_formulation} is a large-scale mixed-integer optimization problem that jointly couples multi-UAV routing and slot-level task offloading. In terms of computational complexity, the UAV routing part is NP-hard since it jointly determines station assignment and visit sequencing under route-consistency, payload, distance, and mission time constraints while maximizing collection value, which amounts to a capacitated value-collecting multi-UAV routing variant. The task offloading part is also NP-hard. To see this, consider a restricted case where all routing-related quantities are fixed, service windows are predetermined, there is only one time slot, communication and energy budgets are set sufficiently large so that they do not constrain the scheduling decisions, and only UAV execution is allowed. This restricted case still subsumes the classical PARTITION problem. PARTITION refers to the problem of deciding whether a set of positive integers can be divided into two subsets with equal sums. Therefore, both subproblems, and thus the full joint formulation, are NP-hard, which motivates our proposed hierarchical DRL approach in \cref{sec:MDP} as a scalable solution method. 

\section{Proposed Agentic AI Framework: LLM with Chain-of-thought}
\label{sec:Proposed_Framework}

\begin{figure}[pos=htbp]  
  \centering
  \includegraphics[width=0.9\textwidth]{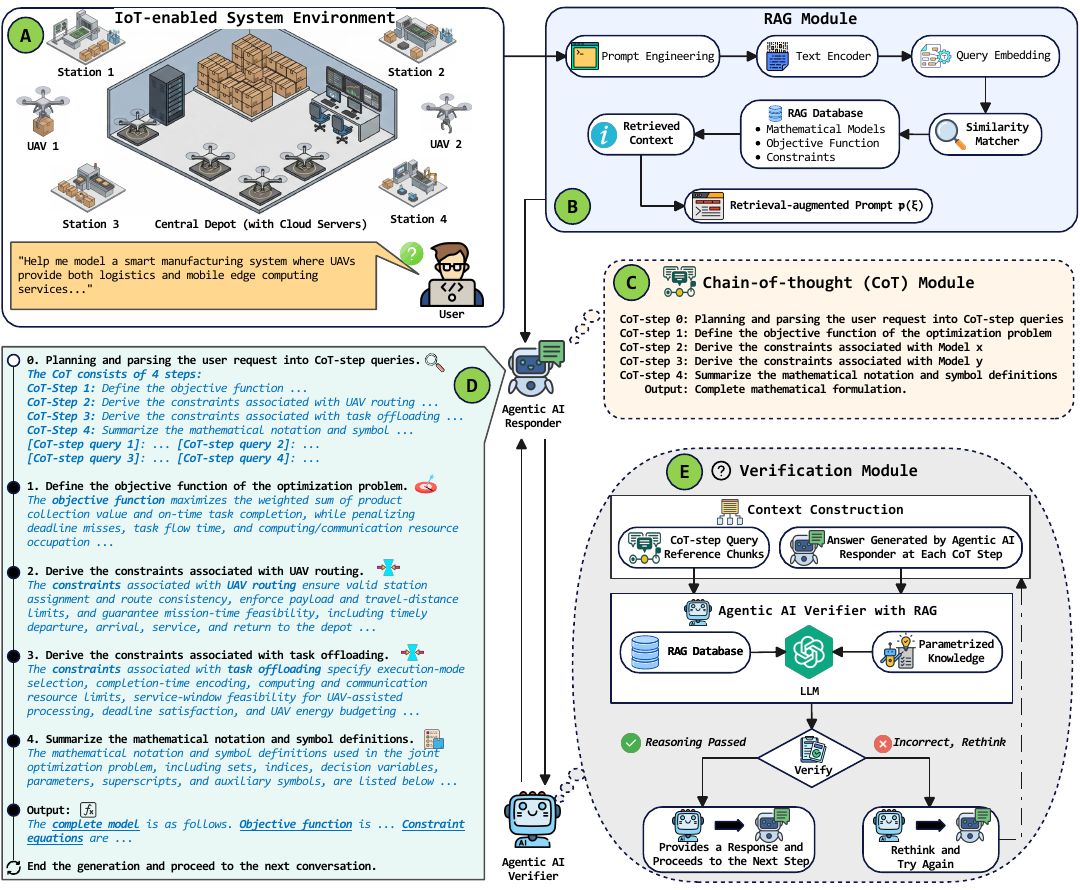}
    \caption{An overview of the proposed agentic AI framework. \textit{Part A} presents the IoT-enabled system environment and the user’s natural-language modeling request. \textit{Part B} illustrates the RAG module for retrieving modeling knowledge and building a retrieval-augmented prompt. \textit{Part C} outlines the CoT module for step-by-step formulation reasoning. \textit{Part D} corresponds to the \textit{Agentic AI Responder} that generates the formulation outputs. \textit{Part E} depicts the verification module, where the \textit{Agentic AI Verifier} checks each CoT-step output and either permits continuation or triggers rethinking.}
  \label{fig:LLM_CoT_flowchart}
\end{figure}

In this section, we introduce the proposed agentic AI framework shown in \cref{fig:LLM_CoT_flowchart}, where the user submits a natural-language modeling request, the request is grounded by the RAG module through retrieval of relevant contextual knowledge, the \textit{Agentic AI Responder} generates the formulation through CoT reasoning, and the \textit{Agentic AI Verifier} validates each reasoning step before the process proceeds. We discuss the retrieval-augmented generation (RAG) processes, and the CoT-based generation-and-verification workflow. 

\subsection{Retrieval-Augmented Generation}

The RAG module \cite{Ruichen2} is used to ground the LLM with task-relevant modeling evidence before mathematical formulation. In this paper, RAG is not used as a general question-answering module. Instead, it supports formulation generation for the hybrid UAV-assisted product collection and MEC scheduling problem. 
Since no public corpus directly matches the considered UAV-assisted CMfg scenario, we construct a domain-specific RAG knowledge base consisting of 11 domain-knowledge Markdown files. These files organize reusable modeling knowledge for UAV routing and MEC task scheduling, including system-modeling principles, notation design, objective construction, resource constraints, deadline modeling, service-window coupling, energy modeling, and system-boundary variants. The resulting knowledge base provides structured modeling evidence that can be retrieved and adapted during mathematical formulation. Additional information on the dataset is provided in \cref{sec:Dataset_Construction}.

The knowledge base is implemented as a formulation-oriented corpus consisting of 11 domain-knowledge Markdown files, whose main contents are summarized in \cref{tab:rag_knowledge_base}. It covers modeling principles related to system modeling, UAV operations, MEC architectures, objective design, notation, UAV routing, task offloading, service-window coupling, deadline handling, and UAV energy budgeting. For exposition, these contents can be grouped into three formulation-related categories. The UAV-routing category covers product collection and UAV mission feasibility. The MEC-service category covers task execution modes, resource allocation, deadline requirements, and UAV energy consumption. The cross-layer coupling category incorporates the relation between UAV service windows and UAV-assisted task processing. These categories show how the knowledge base covers both logistics decisions and computational task scheduling decisions. The corpus also includes system-boundary variants and alternative modeling assumptions, which provide additional domain knowledge for adapting the formulation to different system settings.

\begin{table}[htbp]
	\centering
	\caption{Main contents of the RAG knowledge base for formulation generation.}
	\label{tab:rag_knowledge_base}
	\begingroup
	\setlength{\tabcolsep}{4pt}
	\renewcommand{\arraystretch}{1.08}
	\renewcommand{\tabularxcolumn}[1]{p{#1}}
	\begin{tabularx}{\textwidth}{
	@{}
	>{\raggedright\arraybackslash}p{0.24\textwidth}
	>{\raggedright\arraybackslash}p{0.36\textwidth}
	>{\raggedright\arraybackslash}X
	@{}
	}
	\toprule
	\textbf{Knowledge item}
	& \textbf{Stored content}
	& \textbf{Use in formulation generation} \\
	\midrule

	System entities
	& UAVs, depot, manufacturing stations, ISDs, computational tasks, and cloud support.
	& Helps the LLM identify the physical and computational components in the CMfg scenario. \\

	Notation records
	& Sets, indices, parameters, decision variables, and superscripts used in the UAV-routing and MEC-service models.
	& Helps maintain notation consistency when generating variables and constraints. \\

	Objective patterns
	& Collection value, on-time completion return, deadline-miss cost, flow-time cost, and resource-occupation cost.
	& Helps the LLM construct an objective function that considers both logistics and MEC services. \\

	UAV-routing constraints
	& Station assignment, route consistency, depot balance, payload budget, distance budget, travel time, return time, and station service duration.
	& Grounds the generation of product-collection and UAV mission-feasibility constraints. \\

	MEC-service constraints
	& Execution-mode selection, task-completion encoding, computing-capacity limits, communication-capacity limits, mode consistency, cumulative service sufficiency, deadline satisfaction, and UAV energy consumption.
	& Grounds the generation of task-processing and resource-allocation constraints. \\

	Cross-layer coupling records
	& Service-window indicators, station-to-ISD mapping, UAV availability, and routing-induced feasibility for UAV-assisted processing.
	& Ensures that task offloading is generated consistently with UAV routes and service windows. \\

	Generic formulation primitives and model variants  
	& Reusable objective and constraint primitives, together with alternative formulations, applicability conditions, boundary cases, and modeling caveats.    
	& Provides generic modeling primitives that the LLM can select, adapt, and combine according to the current user request. \\ 

	\bottomrule
	\end{tabularx}
	\endgroup
\end{table}
    
The knowledge base is prepared and indexed as follows. First, the corpus consists of 11 domain-knowledge Markdown files that cover system modeling, notation design, objective construction, UAV routing, MEC task offloading, computing and communication capacities, task completion and deadlines, service-window coupling, UAV energy modeling, and system-boundary variants. These files provide reusable modeling principles, generic formulation primitives, applicability conditions, alternatives, and modeling caveats. Second, the knowledge files use generic local notation where needed, while the Responder establishes a consistent notation system according to the current user request and previously verified formulation context. Third, the Markdown files are segmented using a recursive text splitter with structural separators based on Markdown heading levels, paragraph breaks, line breaks, and spaces. The text splitter is configured with a chunk size of 1800 characters and a chunk overlap of 300 characters. The resulting chunks are then indexed for semantic retrieval. 

Let $\xi$ denote a user modeling request, and let $\mathcal{H}=\{h_1,h_2,\dots,h_N\}$ denote the chunked knowledge base. A text encoder $\Psi_{\mathrm{enc}}(\cdot)$ maps the user request and each chunk into a shared embedding space \cite{Ref_IV_RAG3}. Conceptually, the cosine-based relevance between $\xi$ and chunk $h_n$ can be written as
\begin{equation}
	s_n(\xi) =
	\frac{
	\Psi_{\mathrm{enc}}(\xi)^{\top}\Psi_{\mathrm{enc}}(h_n)
	}{
	\|\Psi_{\mathrm{enc}}(\xi)\|_2
	\|\Psi_{\mathrm{enc}}(h_n)\|_2
	},
	\quad n=1,\dots,N.
\end{equation}
Given a retrieval budget $K_{\mathrm{ret}}$, the RAG module retrieves the top-ranked chunks according to the cosine-based relevance:
\begin{equation}
\mathcal{R}_{\mathrm{rag}}(\xi)
=
\left\{
h_n \mid
n \in
\operatorname*{TopK}_{K_{\mathrm{ret}}}
\left(\{s_i(\xi)\}_{i=1}^{N}\right)
\right\}.
\end{equation}
Here, the $\operatorname*{TopK}$ operator returns the indices of the $K_{\mathrm{ret}}$ largest relevance scores, and $\mathcal{R}_{\mathrm{rag}}(\xi)$ denotes the retrieved evidence set for the current modeling request. The retrieval step uses cosine-based similarity \cite{Cosine_Similarity_update}, and the retrieval budget is configured as four chunks for each query. 
During the CoT-based formulation process, the Responder retrieves relevant evidence for the current step query and uses it to construct the corresponding formulation response. The Verifier uses an independent retriever with the same similarity-search configuration to obtain generic modeling evidence for checking the response.
Thus, RAG is integrated into the step-wise formulation process, rather than being a separate standalone question-answering component.

Overall, the RAG module provides domain evidence for CoT-based formulation. It helps the responder identify relevant entities, select suitable formulation elements, and preserve the coupling between UAV routing and MEC task processing. In particular, the retrieved evidence helps indicate to the LLM that UAV-assisted task processing is feasible only when a UAV is serving the station that hosts the corresponding ISD. Therefore, RAG helps reduce unsupported generation while keeping the generated formulation aligned with the UAV-assisted CMfg scenario.

\subsection{Chain-of-Thought and Verification}
After retrieval grounds the LLM with domain-relevant evidence, the CoT module reasons over the retrieved information and organizes it for downstream mathematical formulation. Recent Long-CoT studies characterize this reasoning paradigm by deep reasoning, extensive exploration, and feasible reflection, which support more intricate and coherent reasoning processes \cite{Ref_F_CoT1}. In \cref{fig:LLM_CoT_flowchart}, this module externalizes the intermediate reasoning that links the problem description to the objective terms and constraint families, thereby making the formulation process more transparent, in line with recent work on interactive reasoning and explicit reasoning-chain manipulation \cite{Ref_F_CoT2}. Recent studies further show that verifier-guided CoT can assess or guide reasoning-step correctness, while verifiability-oriented evaluation can assess reasoning quality beyond final-answer accuracy \cite{Ref_F_CoT3,Ref_F_CoT4}. The CoT-based formulation step is particularly important in the CMfg scenario considered here, where UAV routing and computational task offloading are tightly coupled through time-varying service availability, communication and computing limits, energy budgets, and task deadlines. 

Given a user request $\xi$ and the retrieval-augmented prompt $\mathfrak{p}(\xi)$, the LLM performs a structured reasoning process over the retrieved evidence. The framework in \cref{fig:LLM_CoT_flowchart} includes two agents: the \textit{Agentic AI Responder} and the \textit{Agentic AI Verifier}. The \textit{Agentic AI Responder} is responsible for reasoning over the user request and providing the final answer. In contrast, the \textit{Agentic AI Verifier} evaluates another agent's responses, deciding whether to approve or reject it. The CoT module of the \textit{Agentic AI Responder} is shown in \textit{Part C} of \cref{fig:LLM_CoT_flowchart}. The CoT module first analyzes the user query and decomposes it into a sequence of CoT-step queries, so that the original request can be solved in a structured, step-by-step manner. Specifically, during the initialization phase, an LLM parses the user query and generates both a CoT-step plan and the corresponding CoT-step queries for the \textit{Agentic AI Responder}. These CoT-step queries define the reasoning order and specify what the responder should address at each stage \cite{Ref_FC_s1}. 

As shown in \textit{Part D} of \cref{fig:LLM_CoT_flowchart}, the user request is decomposed into four main steps, namely objective function formulation, UAV routing constraints derivation, task offloading constraints derivation, and mathematical notation summarization. During execution, the \textit{Agentic AI Responder} answers the CoT-step queries sequentially \cite{Ref_FC_s2}. For example, in CoT-step 1, the query on defining the objective function is first sent to the responder, which then produces an initial response. This response is subsequently forwarded to the \textit{Agentic AI Verifier} for assessment through the verification module. If the response does not pass verification, the verifier returns feedback requiring the responder to rethink and revise the current answer. In contrast, if the response is verified as correct, the workflow proceeds to the next CoT-step query \cite{Ref_FC_s3}. The workflow then proceeds in the same manner for the remaining CoT steps. CoT-step 2 derives the constraints associated with UAV routing. CoT-step 3 focuses on the constraints for task offloading. CoT-step 4 summarizes the mathematical notation and symbol definitions used in the formulation. Depending on the size and complexity of each model, CoT-step 2 and CoT-step 3 can be further divided into multiple CoT steps. For example, the task-offloading constraints can be divided into three parts: execution-mode, completion-time, and deadline constraints; computing and communication resource constraints; and service-window and UAV-energy constraints. 

For each CoT step, the \textit{Agentic AI Responder} initially formulates a response to the specific CoT-step query; subsequently, the \textit{Agentic AI Verifier} assesses the generated response prior to the continuation of the reasoning process \cite{Ref_FC_s3}. This sequential reasoning-and-verification methodology ensures the system's stepwise completion of the entire CoT workflow, with the final output being generated only subsequent to the successful verification of all CoT-step responses. Upon the successful verification of all CoT-step responses, the system consolidates the verified intermediate outputs into a comprehensive final response for the user. Consequently, from the user's perspective, the overall workflow can be viewed as a direct mapping from the input query to a complete mathematical model. The intermediate reasoning, decomposition, and verification procedures are handled internally by the CoT module in \textit{Part C} and the verification module in \textit{Part E}. Together, these two modules realize an automated multi-agent reasoning-and-verification process between the \textit{Agentic AI Responder} and the \textit{Agentic AI Verifier}.

\textit{Part E} of \cref{fig:LLM_CoT_flowchart} further presents the verification module of the \textit{Agentic AI Verifier}. For each CoT-step query, the relevant reference chunks are obtained from the RAG database and combined with the corresponding response generated by the \textit{Agentic AI Responder} to construct the verification context \cite{Ref_FC_s4}. Based on this context, together with parameterized knowledge, the verifier determines whether the responder’s answer at the current CoT step is semantically correct and logically consistent. If the answer passes verification, the workflow continues to the next CoT step; otherwise, the verifier instructs the responder to rethink and try again until either a satisfactory response is obtained or the maximum number of allowed retries is reached, at which point the verification loop is terminated to avoid an unbounded iteration process. 

It should be noted that the main CoT-step 1 to CoT-step 4 design in \cref{fig:LLM_CoT_flowchart} follows a common CoT process, in which a complex reasoning task is decomposed into intermediate reasoning steps and solved progressively. This sequential design is consistent with prior CoT studies that guide LLMs through step-by-step, incremental, and sequential reasoning for complex tasks \cite{Typical_CoT1,Typical_CoT2}. However, the current verification procedure mainly performs step-level verification within this sequential CoT-style workflow. Therefore, it does not yet include an implemented global backtracking mechanism across different CoT steps. This is a limitation of the proposed framework. Recent CoT studies have also shown that errors in intermediate reasoning steps may affect the final answer, and that verification, supervision, or self-correction mechanisms can improve the reliability of intermediate reasoning \cite{Typical_CoT3}. A future extension is to incorporate a global consistency-checking mechanism into the verification module. Specifically, the verified outputs of all CoT steps could be stored in a global formulation record that tracks decision variables, parameters, indices, objective terms, constraint families, and cross-step dependencies. The verifier would then examine whether later constraints are consistent with the variables, symbols, and definitions introduced in earlier steps. If an inconsistency is detected, the workflow would backtrack to the responsible earlier step, revise its output, and update the dependent subsequent steps. Such a global verification-and-backtracking mechanism would further reduce cross-step inconsistencies and improve the reliability of agentic-AI-assisted mathematical formulation.

\section{Proposed Hierarchical DRL Approach}
\label{sec:MDP}

Following the decomposition principle in \cite{TwoLayerDRL}, we newly design a two-layer PPO framework for the joint UAV-routing and MEC-scheduling problem. Unlike \cite{TwoLayerDRL}, which decomposes multi-UAV task scheduling into assigning pre-existing mission tasks to UAVs and planning each UAV route, our framework separates routing-induced service availability from time-slotted MEC scheduling. The upper layer jointly determines station assignment and visiting order under payload, flight-distance, mission-time, and energy constraints. Accordingly, the lower layer is redesigned as a time-slotted MEC scheduling layer driven by the service windows produced by the upper-layer routes. It decides per-slot execution modes and compute/communication allocation for local, UAV, and cloud-via-UAV processing under capacity and deadline constraints. The key modification is service-window coupling: upper-layer routes generate UAV-station service windows and residual energy headroom, which define the feasibility of UAV-assisted offloading and resource allocation in the lower layer. Throughout this paper, terrestrial MEC servers are referred to as the \enquote{cloud}.

\subsection{Hierarchical Framework Overview}

The proposed hierarchical framework consists of two sequential PPO-based DRL layers. The upper layer solves the multi-UAV routing problem as a centralized MDP to determine visiting sequences and station assignments under energy, capacity, and flight distance constraints. Conditioned on the upper-layer routes, the lower layer optimizes per-slot task execution by jointly deciding offloading destinations and communication/computing resource allocation under service-window, capacity, and deadline constraints. To couple the two layers operationally, the upper layer provides the lower layer with each UAV's service window and remaining energy headroom. Each service window gives the arrival and departure times of a UAV at a station. In the sequential training procedure, the upper layer is trained first. Its best-performing policy is then fixed and used to provide routing-induced inputs for lower-layer training. Therefore, the training information flow is mainly from the routing layer to the MEC scheduling layer. In this implementation, lower-layer MEC scheduling performance is not fed back to update the upper-layer routing policy; incorporating such feedback for upper-layer route refinement is left as an important direction for future work.

\subsection{Upper-layer DRL: MDP Design}

The upper layer governs the coarse-timescale routing decisions of $U$ UAVs tasked with visiting $M$ stations within a mission horizon of duration $T_{\mathrm{mission}}$. Each upper-layer training cycle consists of multiple complete routing episodes, where each episode comprises a sequence of discrete \emph{upper steps}. During an upper step, routing decisions are taken, UAV states evolve according to physical constraints, and rewards are accumulated.

\subsubsection{Upper-layer MDP}

The routing problem at the upper layer is modeled as a finite-horizon Markov decision process (MDP) with the following components 
\[
\mathcal{M}^{\mathrm{up}}
= \bigl(\mathcal{S^{\mathrm{up}}},\mathcal{A^{\mathrm{up}}},\mathcal{R}^{\mathrm{up}},\mathcal{\gamma}^{\mathrm{up}}\bigr),
\]
where $\mathcal{S^{\mathrm{up}}}$, $\mathcal{A^{\mathrm{up}}}$, $\mathcal{R}^{\mathrm{up}}$, and $\mathcal{\gamma}^{\mathrm{up}}$ are the state space, action space, reward function, and discount factor of upper-layer DRL, respectively.

\subsubsection{State space}
At each upper step $t_a$, the state $s_{t_a}^{up}\in\mathcal{S}^{\mathrm{up}}$ is a continuous vector that summarizes four types of information: (i) per-UAV status, including location, remaining energy, payload, and flight-distance budget; (ii) route-progress information, including normalized route length, completion status, and the set of stations already served by each UAV;  (iii) per-station attributes, including service status, normalized value and payload weights, and local service characteristics; and (iv) mission-level context, including normalized mission time, the fraction of completed collections, and UAV--station distance information for feasibility and planning.

\subsubsection{Action space}

At each upper step, the learned upper-layer policy selects a joint action $a_{t_{a}}^{up}$ defined as: 
\begin{equation}
a_{t_{a}}^{up}=\bigl(a^{\mathrm{prio}}_{t_{a}},\; a^{\mathrm{route}}_{t_{a}}\bigr)\in\mathcal{A^{\mathrm{up}}}, 
\end{equation}
where $a^{\mathrm{prio}}_{t_{a}}$ refers to the priority score \cite{priority_scores1,priority_scores2}, and is a real-valued vector whose dimension equals the number of UAVs and whose ranking determines the order in which the UAVs apply their routing decisions during the current upper step. $a^{\mathrm{route}}_{t_{a}}$ is the routing actions \cite{Routing_action1}. For each UAV $u$, the learned upper-layer policy selects one discrete routing symbol from 
\begin{equation}
a_{t_{a}}^{route} (u) \in \{\text{stay},\; \text{add}(1),\dots,\text{add}(M),\; \text{complete}\},
\end{equation}
where \textit{$\text{stay}$} keeps the UAV at its current location. \textit{$\text{add}(m)$} instructs the UAV to visit station $m$ until $M$ stations, while \textit{$\text{complete}$} returns the UAV to the depot and closes its current route. The routing decisions and priority scores constitute the actionable degrees of freedom. Feasibility masks restrict the learned upper-layer policy to physically admissible routing symbols. An action such as \textit{$\text{add}(m)$} becomes infeasible if its flight distance, return distance, payload, minimum service time, or remaining energy would violate safety margins.

\subsubsection{Reward function}

At upper step $t_{a}$, conditioned on the observed state $s_{t_{a}}^{up}$, the learned upper-layer policy outputs action $a_{t_{a}}^{up}$, and the environment transitions to $s_{t_{a}+1}^{up}$, where $t_{a}\in\{0,1,\dots,T^{up}-1\}$, and $T^{up}$ is the number of executed upper steps in the episode.
The per-step reward, $r_{t_{a}}^{up}$, is defined as:
\begin{equation}
\label{eq:upper-step-reward-merged-v5113}
\begin{aligned}
r_{t_{a}}^{up} = 
& \sum_{m \in \mathcal{C}_{t_{a}}} v_m 
\;+\;
b_{\mathrm{cov}}
\;+\;
b_{\mathrm{bal}}
\;+\;
b_{\mathrm{del}} - \kappa_{\mathrm{con}}
\; +\;\psi_{\mathrm{end}}^{up}.
\end{aligned}
\end{equation}
The term $\sum_{m\in\mathcal{C}_{t_a}} v_m$ is the collection value obtained during the transition, where $\mathcal{C}_{t_a}$ denotes the set of stations newly collected at upper step $t_a$, and $v_m$ is the product value of station $m$. The term $b_{\mathrm{cov}}$ is a shaping reward that encourages increasing the coverage ratio of collected stations, $b_{\mathrm{bal}}$ encourages balanced route workloads across UAVs, and $b_{\mathrm{del}}$ rewards route termination according to two factors: the collected route value and the ratio between the collected route value and the closed-tour distance. The term $\kappa_{\mathrm{con}}$ is an aggregated penalty that discourages infeasible decisions, including violations of energy, payload, remaining-distance, duplicate-visit, movement-time, service-time, and mission-completion constraints. The last term $\psi_{\mathrm{end}}^{up}$ is the terminal-condition term, activated only at the final upper step. It adjusts the final reward according to the terminal outcome by granting a success bonus or imposing a penalty based on the number of unserved stations.

\subsection{Information Flow from the Upper Layer to the Lower Layer}

Each completed upper-layer routing episode provides two types of information for subsequent lower-layer training cycles: the feasible service windows extracted for each UAV-station pair from the corresponding mission-time intervals, and the residual energy headroom of all UAVs at the end of the episode, which is used to scale certain lower-layer resource and capacity conditions.

\subsection{Lower-layer DRL: MDP Design}
\label{sec:lowerMDP_Design}

In each time slot, the lower-layer DRL scheduler performs task scheduling and resource allocation for computation tasks generated by multiple ISDs. Each task can be executed locally, offloaded to a UAV (edge computing), or offloaded to the cloud via a selected UAV. Tasks may span multiple slots. The lower-layer DRL scheduler seeks to accelerate task progress and completion, reduce backlog, and avoid deadline misses, subject to per-slot compute/communication capacity constraints and time-varying UAV service-window constraints determined by the upper layer.

At the beginning of the time slot, the environment collects all unfinished tasks, including (i) waiting tasks and (ii) active tasks that have started but are not yet completed. These tasks are consolidated into a global queue and prioritized according to urgency using the earliest-deadline-first (EDF) policy. The lower-layer DRL scheduler then scans the global queue from most urgent to least urgent and selects up to a preset maximum number of tasks to process in this slot. For each selected task, it first chooses exactly one execution location (local / a specific UAV / cloud via a specific UAV) and then assigns discrete compute and communication resource levels to advance task processing. The aggregated allocations across all selected tasks must remain feasible under the capacity and service-window constraints. 

\subsubsection{Lower-layer MDP}

The lower-layer scheduling problem is modeled as an MDP defined by the tuple:

\[
\mathcal{M}^{\mathrm{lo}}
= \bigl(\mathcal{S^{\mathrm{lo}}},\mathcal{A^{\mathrm{lo}}},\mathcal{R}^{\mathrm{lo}},\mathcal{\gamma}^{\mathrm{lo}}\bigr),
\]
where $\mathcal{S^{\mathrm{lo}}}$, $\mathcal{A^{\mathrm{lo}}}$, $\mathcal{R}^{\mathrm{lo}}$, and $\mathcal{\gamma}^{\mathrm{lo}}$ are the state space, action space, reward function, and discount factor of lower-layer DRL, respectively.

\subsubsection{State space}
The lower-layer state $s_{t_b}^{lo}\in\mathcal{S}^{lo}$ is observed at the beginning of slot $t_b$ and summarizes local queues, UAV-side resources, and current service feasibility through six groups of normalized features: (i) per-ISD local statistics, including local compute capability and headroom, queue and active-local workload summaries, and the minimum remaining time-to-deadline; (ii) per-UAV statistics, including energy headroom, compute and communication capacities, and workload summaries of tasks currently executed on the UAV or relayed to the cloud via that UAV; (iii) global system scalars, such as the current and remaining time, aggregated counts of waiting and active tasks across processing modes, and a summary statistic of UAV energy; (iv) per UAV--ISD active offload status, represented by normalized aggregates of remaining workload and communication bits for active tasks from ISD $k$ associated with UAV $u$; (v) a UAV--ISD service-window availability mask indicating whether UAV $u$ can serve ISD $k$ at time $t_b$; and (vi) a fixed-length, zero-padded snapshot of the top-$N_q$ unfinished tasks ordered by serviceability and urgency, with per-task attributes including origin, status, execution mode and UAV index when applicable, remaining time-to-deadline, and remaining compute and communication requirements.

\subsubsection{Action space}

At the beginning of each slot $t_b$, the learned lower-layer policy outputs a joint action defined over a cached global-queue snapshot $Q_{t_b}$. This joint action specifies the service decisions for the first $K_g$ queue positions. $K_g$ is a fixed truncation size and only the first $\min\{K_g,|Q_{t_b}|\}$ entries take effect. The top-$K_g$ is different from the top-$N_q$ queue snapshot used in the state. $N_q$ denotes the number of highest-priority unfinished tasks encoded in the state. $K_g$ denotes the number of highest-priority queue positions on which the learned lower-layer policy outputs per-slot decisions.
The action is given by
\begin{equation}
A_{t_b}^{lo}
=
\Bigl\{
\bigl(\ell_{t_b}^{(j)},\,\tilde{\alpha}_{t_b}^{(j)},\,\tilde{\beta}_{t_b}^{(j)}\bigr)
\Bigr\}_{j=1}^{K_g},
\end{equation}
where the $j$-th tuple gives the processing decision and resource-allocation scalars for the task at position $j$ in $Q_{t_b}$.
The discrete variable $\ell_{t_b}^{(j)}$ specifies a processing choice for the queue item at position $j$. It is selected from the semantic action set:
\begin{equation}
\ell_{t_b}^{(j)} \in \{\texttt{SKIP},\ \text{local}\} \cup \{\text{UAV}(u),\ \text{cloud-via-UAV}(u)\}_{u=1}^{U},
\end{equation}
where \texttt{SKIP} denotes no new dispatch, while the other options correspond to local execution, UAV execution, or cloud execution via UAV $u$. The \texttt{SKIP} option allows the learned lower-layer policy to serve fewer tasks in a slot when needed to respect resource limits and prioritize more urgent or more serviceable tasks. 

The scalars $\tilde{\alpha}_{t_b}^{(j)}, \tilde{\beta}_{t_b}^{(j)} \in [0,1]$ denote normalized compute and communication allocations, which are quantized by the environment into discrete levels. The continuous sampling of $\tilde{\alpha}_{t_b}^{(j)}$ and $\tilde{\beta}_{t_b}^{(j)}$ is adopted to avoid a high-dimensional multi-discrete action space, while quantization ensures feasibility under level-based resource constraints. For local execution, communication allocation is ignored. 
For active tasks already bound to a processing mode and UAV index, $\ell_{t_b}^{(j)}$ is restricted to either \texttt{SKIP} or the previously assigned processing option. If selected, the allocation is updated for the current slot; otherwise, previous allocations persist.

\subsubsection{Reward function}
\label{Test51_13_rewards}
The step reward $r^{\mathrm{lo}}_{t_b}$ is designed to align with the objectives:
(i) finish tasks as early as possible, 
(ii) strictly avoid deadline failures, and 
(iii) discourage avoidable idling and excessive (always-max) resource occupation. 
A dense-and-aligned reward is written as: 
\begin{equation}\label{eqn_lower_DRL_rewards_test51_13}
\begin{aligned}
r^{\mathrm{lo}}_{t_b} = b_{\mathrm{complete}}
+ b_{\mathrm{dispatch}}
+ b_{\mathrm{progress}} 
+ \Phi_{\mathrm{B}} + \Phi_{\mathrm{U}} - \kappa_{\mathrm{deadline}} - \kappa_{\mathrm{invalid}} - \kappa_{\mathrm{idle}}
- \kappa_{\mathrm{alloc}}
- \kappa_{\mathrm{living}} + \psi_{\mathrm{end}}^{lo}. 
\end{aligned}
\end{equation}
At slot $t_b$, the lower-layer reward $r_{t_b}^{\mathrm{lo}}$ combines positive incentives, penalties, and an end-of-episode settlement. Specifically, $b_{\mathrm{complete}}$ rewards task completions in slot $t_b$, $b_{\mathrm{dispatch}}$ assigns a bonus or cost to each waiting task dispatched in that slot according to its selected execution destination, e.g., local, UAV, or cloud via a UAV. $b_{\mathrm{progress}}$ provides a dense reward for effective task advancement during the slot. The shaping terms $\Phi_{\mathrm{B}}$ and $\Phi_{\mathrm{U}}$ further encourage backlog reduction and urgency reduction, respectively, by rewarding transitions that reduce the waiting-task backlog and the urgency of unfinished tasks. In contrast, $\kappa_{\mathrm{deadline}}$ penalizes deadline violations, $\kappa_{\mathrm{invalid}}$ penalizes infeasible decisions that violate service-window or resource-capacity constraints, and $\kappa_{\mathrm{idle}}$ penalizes avoidable \enquote{no-dispatch} behavior when feasible waiting tasks exist. In addition, $\kappa_{\mathrm{alloc}}$ discourages excessive resource usage by penalizing the aggregate normalized allocation levels of computation and communication resources, and $\kappa_{\mathrm{living}}$ imposes a persistent penalty on the remaining unfinished load after the current transition so as to maintain continuous pressure for prompt task clearance. Finally, $\psi_{\mathrm{end}}^{lo}$ denotes the terminal contribution applied at the end of the episode, which penalizes unfinished tasks remaining at termination, imposes a stronger penalty on overdue tasks, and may also provide a sparse bonus when the system is cleared early after the final task arrival. 

The compact notation in \cref{eqn_lower_DRL_rewards_test51_13} groups several weighted sub-terms into interpretable reward components. We discuss them in details as follows. 
The completion reward is defined as
\begin{equation}
b_{\mathrm{complete}}
=
\alpha_{\mathrm{c}} N^{\mathrm{comp}}_{t_b}, 
\end{equation}
where $N^{\mathrm{comp}}_{t_b}$ is the number of tasks completed in slot $t_b$, and $\alpha_{\mathrm{c}}$ is the per-task completion reward.
The dispatch term is defined as
\begin{equation}
b_{\mathrm{dispatch}}
=
\beta_{\mathrm{uav}} N^{\mathrm{uav}}_{t_b}
+
\beta_{\mathrm{loc}} N^{\mathrm{loc}}_{t_b}
+
\beta_{\mathrm{cld}} N^{\mathrm{cld}}_{t_b}, 
\end{equation}
where $N^{\mathrm{uav}}_{t_b}$, $N^{\mathrm{loc}}_{t_b}$, and $N^{\mathrm{cld}}_{t_b}$ are the numbers of newly dispatched tasks assigned to UAV execution, local execution, and cloud execution via a UAV, respectively. The coefficients $\beta_{\mathrm{uav}}$, $\beta_{\mathrm{loc}}$, and $\beta_{\mathrm{cld}}$ specify the reward or cost associated with each dispatch destination. 
The progress reward is defined as
\begin{equation}
b_{\mathrm{progress}}
=
\alpha_{\mathrm{prog}}
\left(
\Delta W^{\mathrm{net}}_{t_b}
+
\frac{\Delta B^{\mathrm{net}}_{t_b}}{\xi_{\mathrm{bit}}}
\right), 
\end{equation}
where $\Delta W^{\mathrm{net}}_{t_b}$ and $\Delta B^{\mathrm{net}}_{t_b}$ denote the net reduction in remaining workload and remaining communication bits during slot $t_b$, respectively, after subtracting workload and communication-bit reductions caused only by deadline expiration. The parameter $\xi_{\mathrm{bit}}$ converts transmitted bits into workload-equivalent progress units, and $\alpha_{\mathrm{prog}}$ scales the dense progress reward. 
The backlog-potential shaping term is defined as
\begin{equation}
\Phi_{\mathrm{B}}
=
\phi_{\mathrm{B}}(s_{t_b}^{\mathrm{lo}})
-
\phi_{\mathrm{B}}(s_{t_b+1}^{\mathrm{lo}}),
\qquad
\phi_{\mathrm{B}}(s)
=
\min\!\left\{
c_{\mathrm{B}}\bar{N}^{\mathrm{wait}}(s),
\phi_{\mathrm{B}}^{\max}
\right\},
\end{equation}
where the subscript $\mathrm{B}$ denotes backlog. $\bar{N}^{\mathrm{wait}}(s)$ is the average number of waiting tasks per ISD in state $s$. The coefficient $c_{\mathrm{B}}$ scales the backlog potential, and $\phi_{\mathrm{B}}^{\max}$ caps this potential. Therefore, $\Phi_{\mathrm{B}}$ is positive when the transition reduces the backlog potential. 
The urgency-potential shaping term is defined as
\begin{equation}
\Phi_{\mathrm{U}}
=
\phi_{\mathrm{U}}(s_{t_b}^{\mathrm{lo}})
-
\phi_{\mathrm{U}}(s_{t_b+1}^{\mathrm{lo}}),
\qquad
\phi_{\mathrm{U}}(s)
=
c_{\mathrm{U}}\bar{\upsilon}(s),
\end{equation}
where the subscript $\mathrm{U}$ denotes urgency. The term $\bar{\upsilon}(s)$ is the average over-threshold urgency of unfinished tasks, where only the urgency portion above $\theta_{\mathrm{U}}^{\mathrm{lo}}$ is counted. The coefficient $c_{\mathrm{U}}$ scales the urgency potential. Therefore, $\Phi_{\mathrm{U}}$ is positive when the transition reduces the urgency potential. 
The deadline-violation penalty is defined as
\begin{equation}
\kappa_{\mathrm{deadline}}
=
\alpha_{\mathrm{d}} N^{\mathrm{dl}}_{t_b},
\end{equation}
where $N^{\mathrm{dl}}_{t_b}$ is the number of deadline violations in slot $t_b$, and $\alpha_{\mathrm{d}}$ is the per-task deadline-miss penalty. 
The infeasible-decision penalty is defined as
\begin{equation}
\kappa_{\mathrm{invalid}}
=
\alpha_{\mathrm{inv}} N^{\mathrm{inv}}_{t_b},
\end{equation}
where $N^{\mathrm{inv}}_{t_b}$ is the number of infeasible decisions in slot $t_b$, and $\alpha_{\mathrm{inv}}$ is the infeasible-decision penalty coefficient. 
The avoidable-idling penalty is defined as
\begin{equation}
\kappa_{\mathrm{idle}}
=
\alpha_{\mathrm{idle}}\rho^{\mathrm{idle}}_{t_b}, 
\end{equation}
where $\rho^{\mathrm{idle}}_{t_b}=1$ if feasible waiting tasks exist but no waiting task is dispatched in slot $t_b$, and $\rho^{\mathrm{idle}}_{t_b}=0$ otherwise. The coefficient $\alpha_{\mathrm{idle}}$ penalizes avoidable no-dispatch behavior. 
The resource-occupation penalty is defined as
\begin{equation}
\kappa_{\mathrm{alloc}}
=
\lambda_{\mathrm{cmp}} R^{\mathrm{cmp}}_{t_b}
+
\lambda_{\mathrm{com}} R^{\mathrm{com}}_{t_b},
\end{equation}
where $R^{\mathrm{cmp}}_{t_b}$ and $R^{\mathrm{com}}_{t_b}$ are defined in \cref{eqn_R_cmp,eqn_R_com} respectively. Thus, $\kappa_{\mathrm{alloc}}$ penalizes normalized occupation rather than raw resource allocation. The coefficients $\lambda_{\mathrm{cmp}}$ and $\lambda_{\mathrm{com}}$ control the relative penalties for computation and communication occupation, respectively. 
The living penalty is defined as 
\begin{equation}
\kappa_{\mathrm{living}}
=
\lambda_{\mathrm{B}}^{\mathrm{liv}}\bar{N}^{\mathrm{wait}}_{t_b+1}
+
\lambda_{\mathrm{U}}^{\mathrm{liv}}\bar{\upsilon}_{t_b+1}
+
\lambda_{\mathrm{A}}^{\mathrm{liv}}\bar{N}^{\mathrm{act}}_{t_b+1}, 
\end{equation}
where $\bar{N}^{\mathrm{wait}}_{t_b+1}$, $\bar{\upsilon}_{t_b+1}$, and $\bar{N}^{\mathrm{act}}_{t_b+1}$ are the average waiting-task count, average urgency level, and average active-task count after the transition, respectively. The coefficients $\lambda_{\mathrm{B}}^{\mathrm{liv}}$, $\lambda_{\mathrm{U}}^{\mathrm{liv}}$, and $\lambda_{\mathrm{A}}^{\mathrm{liv}}$ determine the persistent pressure on the remaining waiting backlog, urgency, and active task load. 
The terminal contribution is computed as
\begin{equation}
\psi_{\mathrm{end}}^{lo}
=
-\alpha_{\mathrm{term}}N^{\mathrm{unfin}}_{\mathrm{end}}
-\alpha_{\mathrm{d}}N^{\mathrm{dl}}_{\mathrm{end}}
+
\alpha_{\mathrm{clear}}
\frac{\left[T_{\mathrm{mission}}-t_{\mathrm{clear}}\right]_+}{T_{\mathrm{mission}}},
\end{equation}
where $N^{\mathrm{unfin}}_{\mathrm{end}}$ is the number of unfinished but non-overdue tasks at termination, $N^{\mathrm{dl}}_{\mathrm{end}}$ is the number of overdue tasks at termination, $t_{\mathrm{clear}}$ is the first slot in which the system becomes empty after the final task arrival, and $[x]_+=\max\{x,0\}$. If the system is not cleared, the early-clear bonus is set to zero. This terminal term separately settles non-overdue unfinished tasks and overdue tasks at the end of the episode. 
The coefficients in the lower-layer reward function are listed in \cref{tab:lower_reward_coefficients}.
\begin{table}[t]
    \centering
    \caption{List of coefficients in the lower-layer reward function.}
    \label{tab:lower_reward_coefficients}
    \begin{tabular}{
    p{0.8cm}
    p{0.7cm}    
    p{4.6cm}
    p{0.8cm}    
    p{1.06cm}   
    p{5.8cm}
    }
    \toprule
    \textbf{Symbol} & \textbf{Value} & \textbf{Description} 
    & \textbf{Symbol} & \textbf{Value} & \textbf{Description} \\
    \midrule
    $\alpha_{\mathrm{c}}$ & $0.15$ & Per-task completion reward
    & $c_{\mathrm{B}}$ & $3.0$ & Backlog-potential scaling \\
    $\alpha_{\mathrm{clear}}$ & $10.0$ & Early-clear terminal bonus scaling
    & $c_{\mathrm{U}}$ & $1.5$ & Urgency-potential scaling \\
    $\alpha_{\mathrm{d}}$ & $0.6$ & Per-task deadline-miss penalty
    & $\lambda_{\mathrm{A}}^{\mathrm{liv}}$ & $0.05$ & Persistent active-task penalty weight \\
    $\alpha_{\mathrm{idle}}$ & $0.15$ & Avoidable no-dispatch penalty
    & $\lambda_{\mathrm{B}}^{\mathrm{liv}}$ & $0.05$ & Persistent waiting-backlog penalty weight \\
    $\alpha_{\mathrm{inv}}$ & $0.3$ & Infeasible-decision penalty
    & $\lambda_{\mathrm{cmp}}$ & $0.8$ & Computation-occupation penalty weight \\
    $\alpha_{\mathrm{prog}}$ & $0.0025$ & Dense progress scaling
    & $\lambda_{\mathrm{com}}$ & $0.5$ & Communication-occupation penalty weight \\
    $\alpha_{\mathrm{term}}$ & $0.6$ & Terminal unfinished-task penalty
    & $\lambda_{\mathrm{U}}^{\mathrm{liv}}$ & $0.02$ & Persistent urgency penalty weight \\
    $\beta_{\mathrm{cld}}$ & $-0.19$ & Cloud-dispatch bias
    & $\phi_{\mathrm{B}}^{\max}$ & $200.0$ & Backlog-potential cap \\
    $\beta_{\mathrm{loc}}$ & $-0.22$ & Local-dispatch bias
    & $\theta_{\mathrm{U}}^{\mathrm{lo}}$ & $0.2$ & Urgency activation threshold \\
    $\beta_{\mathrm{uav}}$ & $0.35$ & UAV-dispatch bonus
    & $\xi_{\mathrm{bit}}$ & $4.0\times 10^{5}$ & Bit-to-work-unit conversion \\
    \bottomrule
    \end{tabular}
\end{table}

\subsection{Complexity Analysis of Upper-- and Lower--Layer DRL}

Let $N_{\mathrm{up}}$ and $N_{\mathrm{lo}}$ denote the numbers of training episodes for the upper and lower layers, respectively, and let $\bar{T}^{\mathrm{up}}$ and $\bar{T}^{\mathrm{lo}}$ denote the average episode lengths. Let $P_{\theta,\phi}^{\mathrm{up}}$ and $P_{\theta,\phi}^{\mathrm{lo}}$ denote the total numbers of trainable parameters in the upper-layer and lower-layer actor--critic networks, respectively. In the upper layer, routing-action evaluation and feasibility-mask construction scale with the number of UAV--station pairs. Thus, the upper-layer training complexity is $\mathcal{O}(N_{\mathrm{up}}\bar{T}^{\mathrm{up}}(UM+P_{\theta,\phi}^{\mathrm{up}}))$, under fixed PPO update settings. In the lower layer, the state contains per-ISD features, per-UAV features, UAV--ISD service-window masks, and a top-$N_q$ queue snapshot, while the policy only acts on the first $K_g$ queue positions. Hence, the lower-layer training complexity is $\mathcal{O}(N_{\mathrm{lo}}\bar{T}^{\mathrm{lo}}(UK+N_q+K_gU+P_{\theta,\phi}^{\mathrm{lo}}))$. A monolithic MDP would need to enumerate joint routing and task-scheduling choices, whose size scales with $(M+2)^U(2U+2)^{K_g}$ before considering resource-allocation levels. Therefore, the proposed hierarchical decomposition keeps the rollout and update costs polynomial in the main system sizes while retaining the service-window information needed by the lower layer. This reduction in complexity comes with a limitation. The method does not optimize a single end-to-end policy over both routing and MEC-scheduling actions, and the lower-layer MEC performance is not used to refine the upper-layer routing policy during training.

\subsection{Training Procedure}
\label{sec:training}

The two-layer DRL framework is trained sequentially. First, the upper layer is trained by PPO for UAV routing, where routing rollouts are collected under feasibility masks and the actor--critic networks are updated whenever the rollout buffer is full. The best-performing upper-layer policy is then fixed to extract the UAV service-window and residual-energy information for the lower layer. Next, the lower layer is trained by PPO for task scheduling and resource allocation under this fixed routing information. In this stage, scheduling rollouts are collected over EDF-based queue snapshots with feasibility masks and quantized allocation decisions, and the actor--critic networks are updated whenever the rollout buffer is full. 

This sequential design is adopted for tractability and training stability. It avoids direct optimization of a monolithic MDP whose joint routing and task-scheduling action space grows rapidly with the numbers of UAVs, stations, and scheduled tasks. However, this design should be interpreted as a tractable hierarchical approximation. The upper layer provides service windows and residual energy to the lower layer, but lower-layer MEC performance is not fed back to update the upper-layer routing policy during training. Therefore, the learned upper-layer policy is mainly optimized for collection-oriented routing feasibility. It does not guarantee optimality for the fully coupled routing-MEC objective, especially when MEC demand varies significantly across stations or over time. However, this design can avoid the non-convergence issue. That is, if the upper layer and lower layer are both iterative, the solution may not converge. Accordingly, extending the current framework to a feedback-aware hierarchical training scheme with convergence-control mechanisms is left as future work.

\section{Performance Evaluation}
\label{sec:Results}

\subsection{Simulation Parameters and Setup}
In this subsection, we describe the scenario settings, agentic AI settings, and DRL settings as follows.

\subsubsection{Scenario Settings}
We consider a cloud-manufacturing scenario with two homogeneous UAVs serving six manufacturing stations, where each station is equipped with one ISD. The mission horizon is set to 313~s. The minimum required service durations are set to 100~s at the first station, 80~s at the second, third, and fourth stations, and 70~s at the fifth and sixth stations. Here, payload-units denote an abstract unit used to measure product weight and UAV payload budget. For MEC scheduling, the lower layer is discretized into 1-s time slots, and work-units are used as an abstract unit to measure computation workload and processing capacity. In each episode, 1920 tasks are randomly generated, with deadlines uniformly distributed from 20~s to 80~s. The task workload is randomly generated around 10 work-units per task, with a typical deviation of about 2 work-units, while the task input size is randomly generated around $10^{5}$ bits per task, with a typical deviation of about $2\times10^{4}$ bits. The other scenario settings are listed in \cref{tab:table1_hyper}.

\begin{table}[pos=htbp]
    \centering
    \caption{Scenario settings}
    \label{tab:table1_hyper}
    \begingroup
    \setlength{\tabcolsep}{4pt}
    \renewcommand{\arraystretch}{1.08}
    \renewcommand{\tabularxcolumn}[1]{p{#1}}
    \begin{tabularx}{\textwidth}{
        @{}
        >{\raggedright\arraybackslash}p{3.5cm}
        >{\raggedright\arraybackslash}X
        >{\raggedright\arraybackslash}p{5.0cm}
        >{\raggedright\arraybackslash}X
        @{}
    }
        \toprule
        \textbf{System Parameter} 
        & \textbf{Value} 
        & \textbf{System Parameter} 
        & \textbf{Value} \\
        \midrule

        UAV flight speed 
        & $10$ m/s 
        & Local computing capacity 
        & $300$ work-units/s \\

        Maximum travel distance 
        & $4800$ m 
        & UAV computing capacity 
        & $400$ work-units/s \\

        UAV payload capacity 
        & $30$ payload-units 
        & Cloud computing capacity 
        & $400$ work-units/s \\

        UAV battery capacity 
        & $150$ Wh 
        & ISD-to-UAV / UAV-to-cloud rate 
        & $500 / 120$ Mb/s \\

        \bottomrule
    \end{tabularx}
    \endgroup
\end{table}

The adopted configuration should be interpreted as a nominal-load, feasibility-oriented scenario. It is used to evaluate whether the hierarchical policy can coordinate routing-induced service availability with task-execution and resource-allocation decisions while satisfying the modeled feasibility constraints. Similar controlled simulation settings, in which UAV-assisted MEC policies are evaluated under representative system configurations while computation offloading and resource-allocation decisions remain subject to system constraints, have also been adopted in the literature \cite{Ref_DRL_design1,Ref_DRL_design2}. It is not intended as a capacity-saturation stress test. Because the computing and communication capacities are relatively large compared with the average task demand, the current results do not quantify performance when these resources become persistently binding. Future work will introduce resource-constrained configurations by jointly varying task-arrival intensity, workload and input-size distributions, deadline ranges, computing capacities, and uplink and backhaul capacities. These evaluations will report average and peak resource utilization, queue backlog, deadline violations, execution-mode allocation, and the frequency of binding resource constraints.

The current evaluation focuses on the end-to-end scheduling outcomes under the complete set of UAV-energy and resource-capacity constraints. Energy consumption and computing and communication occupation are incorporated into the feasibility conditions and reward design, while the reported metrics emphasize product collection, task-deadline satisfaction, and training performance, which directly reflect the main operational objectives of the proposed framework. A detailed resource-accounting analysis requires a unified evaluation protocol because flight, hovering, onboard computing, uplink transmission, and backhaul transmission involve different physical quantities and scales. Similarly, local, UAV, and cloud computing resources and the two communication links have different capacities and availability conditions. Directly aggregating these heterogeneous quantities may therefore provide an incomplete interpretation. Future work will conduct a dedicated resource-accounting evaluation that separately reports the UAV-energy components, energy-budget utilization, and normalized computation and communication utilization.

\subsubsection{Agentic AI Settings}

The agentic AIs are implemented as follows. First, the expertise knowledge and retrieval queries are encoded using OpenAIEmbeddings with \emph{text-embedding-ada-002} \cite{text_embedding_ada_002}. Next, a local RAG pipeline is constructed by indexing the expertise corpus into a Chroma vector store \cite{Chroma} and exposing similarity-based retrievers as retrieval tools. The planning, response-generation, and verification components all employ GPT-5.6 Luna, with the reasoning effort set to \emph{low}. However, these components operate independently. The planning component is a standalone LLM call rather than an agent. It analyzes the user prompt and decomposes it into a sequence of CoT-step queries. The resulting queries are then processed by two independent agents: a Responder Agent and a Verifier Agent. Although both agents use the same GPT-5.6 Luna model, they have separate instructions, retrieval tools, and short-term conversational memories. Based on the Standard API pricing snapshot recorded on 14 July 2026, the input, cached-input, and output costs of GPT-5.6 Luna are USD~1.00, USD~0.10, and USD~6.00 per million tokens, respectively \cite{gpt_5dot6_Luna}. Short-term conversational memory is maintained separately for the Responder and Verifier using LangGraph checkpointing \cite{LangGraph} within the LangChain-based agent framework \cite{LangChain}. All results reported in \cref{tab:verifier_reliability,tab:multiple_run_results,tab:formulation_quality_ablation} were produced using GPT-5.6 Luna with low reasoning effort.

\subsubsection{DRL Settings}

We adopt a hierarchical DRL framework for the considered hybrid logistics--MEC scheduling problem. In both compared schemes, the upper layer uses PPO to optimize UAV routing, while the lower layer is used to optimize task execution and resource allocation. To evaluate the effectiveness of our proposed approach, we compare it with a baseline method using advantage actor-critic (A2C). The baseline scheme employs PPO in the upper layer and A2C in the lower layer, whereas the proposed scheme employs PPO in both the upper and lower layers. For both layers, the actor and critic networks use two hidden layers with 256 neurons per layer. ReLU is adopted as the activation function, and Adam is used for network optimization. The detailed hyperparameter settings of the upper-layer PPO (Up-PPO), lower-layer PPO (Low-PPO), and lower-layer A2C (Low-A2C) are summarized in \Cref{tab:drl_hyper_compact}.

\begin{table}[pos=htbp]
    \centering
    \caption{DRL hyperparameter settings}
    \label{tab:drl_hyper_compact}
    \begingroup
    \setlength{\tabcolsep}{4pt}
    \renewcommand{\arraystretch}{1.08}
    \renewcommand{\tabularxcolumn}[1]{p{#1}}
    \begin{tabularx}{\textwidth}{
        @{}
        >{\raggedright\arraybackslash}p{8.20cm}
        >{\centering\arraybackslash}X
        >{\centering\arraybackslash}X
        >{\centering\arraybackslash}X
        @{}
    }
        \toprule
        \textbf{System Parameter} 
        & \textbf{Up-PPO} 
        & \textbf{Low-PPO} 
        & \textbf{Low-A2C} \\
        \midrule

        Learning rate (actor)   
        & $2\times10^{-4}$ 
        & $7\times10^{-5}$ 
        & $1\times10^{-5}$ \\

        Learning rate (critic)  
        & $3\times10^{-4}$ 
        & $1\times10^{-4}$ 
        & $5\times10^{-5}$ \\

        Discount factor         
        & $0.99$           
        & $0.996$          
        & $0.996$ \\

        Generalized advantage estimation / trace parameter   
        & $0.92$           
        & $0.95$           
        & $0.95$ \\

        Clipping parameter      
        & $0.15$           
        & $0.05$           
        & -- \\

        Entropy coefficient     
        & $0.004$          
        & $0.001$          
        & $0.001$ \\

        Rollout length          
        & $96$             
        & $512$            
        & $64$ \\

        Batch size              
        & $32$             
        & $512$            
        & -- \\

        Training episodes       
        & $2000$           
        & $2000$           
        & $2000$ \\

        Number of hidden layers 
        & $2$              
        & $2$              
        & $2$ \\

        Hidden layer size       
        & $256$            
        & $256$            
        & $256$ \\

        Activation function     
        & ReLU             
        & ReLU             
        & ReLU \\

        Deep neural network optimizer           
        & Adam             
        & Adam             
        & Adam \\

        \bottomrule
    \end{tabularx}
    \endgroup
\end{table}

\subsection{Effectiveness of the Agentic AI}

\begin{figure}[pos=htbp]
  \centering
  \includegraphics[width=0.9\textwidth]{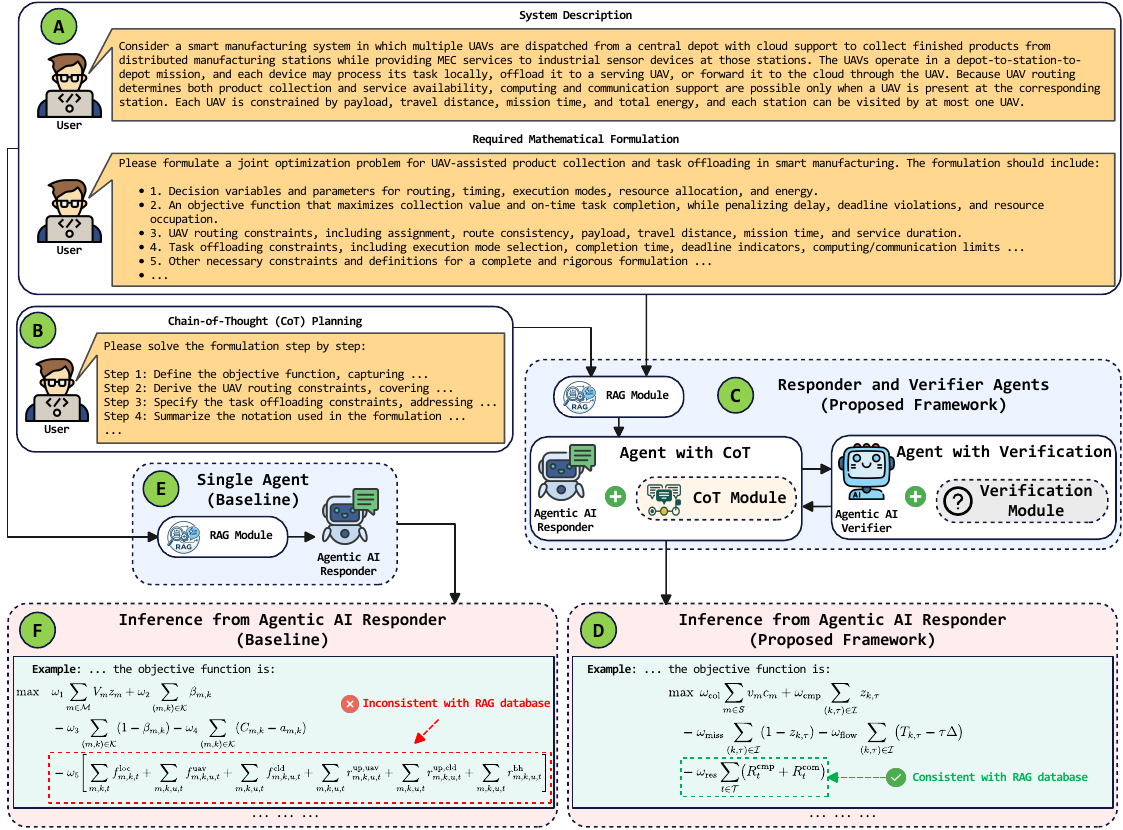}
    \caption{Results from the proposed agentic AI framework. \textit{Part A} presents an example of a user's system description and mathematical formulation request. \textit{Part B} illustrates the CoT planning from the user. \textit{Part C} outlines the conversation between the responder and verifier agents. \textit{Part D} lists the examples from the final answer of \textit{Agentic AI Responder} with CoT module and verification module. \textit{Part E} depicts a single agent framework. \textit{Part F} lists the examples from the final answer of \textit{Agentic AI Responder} without CoT module and verification module.}
  \label{fig:LLM_CoT_Results}
\end{figure}

\Cref{fig:LLM_CoT_Results} presents the results of the proposed agentic AI framework, comprising the \textit{Agentic AI Responder} and \textit{Agentic AI Verifier}, along with a single-agent baseline that includes only the \textit{Agentic AI Responder} without CoT and verification modules. In \textit{Part A} of \Cref{fig:LLM_CoT_Results}, the user first provides the prompts, including the system description and the required mathematical formulation. A CoT planning prompt, shown in \textit{Part B}, is then sent to the proposed framework in \textit{Part C}. Finally, the \textit{Agentic AI Responder}, equipped with both CoT and verification modules, generates the complete response, of which only a partial example is presented in \textit{Part D}. In addition, a single-agent baseline is shown in \textit{Part E}. It consists only of the Agentic AI Responder equipped with RAG and does not include either the CoT module or the verification module. \textit{Part F} presents several examples extracted from its complete response. 

In \Cref{fig:LLM_CoT_Results}, \textit{Part D} and \textit{Part F} present examples generated by the proposed agentic AI framework and the baseline, respectively. The main difference between them lies in their semantic fidelity to the RAG database. The proposed framework remains closer to the original modeling logic by preserving the intended meanings of the modeled terms, whereas the baseline tends to reformulate them into a different optimization-oriented representation. Therefore, the distinction is not merely syntactic, but conceptual. For example, the semantic difference is reflected in what the resource-related penalty is intended to measure. In the RAG database, the penalty is defined through the abstract occupation variables \(R_t^{\mathrm{cmp}}\) and \(R_t^{\mathrm{com}}\), which represent normalized computing and communication occupation. The proposed framework preserves this same interpretation, so the penalty still refers to an aggregated and normalized notion of overall resource occupation. By contrast, the baseline rewrites this penalty as a direct sum of computing and communication allocation variables. As a result, it changes the penalty from normalized resource occupation to raw resource usage. This is a semantic change, not merely a notational difference.

\subsubsection{Verifier Reliability}

To validate the reliability of the verifier, we ran the complete workflow of the proposed agentic AI framework six times and evaluated the verifier using the true acceptance (TA), false acceptance (FA), false rejection (FR), true rejection (TR), false acceptance rate (FAR), and false rejection rate (FRR), with the results summarized in \Cref{tab:verifier_reliability}. 
Let $N_{\mathrm{run}}$ denote the total number of repeated runs, and let $ r \in\{1,\ldots, N_{\mathrm{run}}\} $ denote the run index. For run $ r $, let $ \mathit{TA}_{r} $, $ \mathit{FA}_{r} $, $ \mathit{FR}_{r} $, and $ \mathit{TR}_{r} $ denote the numbers of TA, FA, FR, and TR, respectively. A valid response is treated as a positive case, while an invalid response is treated as a negative case. The number of evaluated cases ($ N_{\mathrm{case},r} $) and the per-run Verifier-reliability metrics are defined as follows:
\begin{equation}
\label{eq:verifier_run_metrics}
\left\{
\begin{alignedat}{2}
 N_{\mathrm{case},r} 
&=
 \mathit{TA}_{r} 
+ \mathit{FA}_{r} 
+ \mathit{FR}_{r} 
+ \mathit{TR}_{r},
\qquad &
 \mathit{FAR}_{r} 
&=
\frac{ \mathit{FA}_{r} }
{ \mathit{FA}_{r} + \mathit{TR}_{r} },
\\[4pt]
 \mathit{Accuracy}_{r} 
&=
\frac{ \mathit{TA}_{r} + \mathit{TR}_{r} }
{ N_{\mathrm{case},r} },
\qquad &
 \mathit{FRR}_{r} 
&=
\frac{ \mathit{FR}_{r} }
{ \mathit{TA}_{r} + \mathit{FR}_{r} },
\end{alignedat}
\right.
\end{equation}
where $ \mathit{FAR}_{r} $ is the proportion of invalid responses that are incorrectly accepted, $ \mathit{FRR}_{r} $ is the proportion of valid responses that are incorrectly rejected, and $ \mathit{Accuracy}_{r} $ is the proportion of all evaluated responses that are classified correctly.

The overall metrics are obtained by pooling the case counts across all repeated runs. The overall number of evaluated cases is $ N_{\mathrm{case},\mathrm{all}} =\sum_{ r =1}^{ N_{\mathrm{run}}} N_{\mathrm{case},r} $. The overall Verifier-reliability metrics are defined as follows:
\begin{equation}
\label{eq:verifier_overall_metrics}
\left\{
\begin{alignedat}{4}
\mathit{TA}_{\mathrm{all}} 
&=
\sum_{ r =1}^{ N_{\mathrm{run}}} \mathit{TA}_{r} , \quad
&
\mathit{FR}_{\mathrm{all}} 
&=
\sum_{ r =1}^{ N_{\mathrm{run}}} \mathit{FR}_{r} , \quad
&
\mathit{FAR}_{\mathrm{all}} 
&=
\frac{ \mathit{FA}_{\mathrm{all}} }
{ \mathit{FA}_{\mathrm{all}} + \mathit{TR}_{\mathrm{all}} }, \quad
&
\mathit{Accuracy}_{\mathrm{all}} 
&=
\frac{ \mathit{TA}_{\mathrm{all}} + \mathit{TR}_{\mathrm{all}} }
{ N_{\mathrm{case},\mathrm{all}} }, 
\\[4pt]
\mathit{FA}_{\mathrm{all}} 
&=
\sum_{ r =1}^{ N_{\mathrm{run}}} \mathit{FA}_{r} , \quad
&
\mathit{TR}_{\mathrm{all}} 
&=
\sum_{ r =1}^{ N_{\mathrm{run}}} \mathit{TR}_{r} , \quad
&
\mathit{FRR}_{\mathrm{all}} 
&=
\frac{ \mathit{FR}_{\mathrm{all}} }
{\mathit{TA}_{\mathrm{all}} + \mathit{FR}_{\mathrm{all}} },
\end{alignedat}
\right.
\end{equation}
where $ \mathit{TA}_{\mathrm{all}} $, $ \mathit{FA}_{\mathrm{all}} $, $ \mathit{FR}_{\mathrm{all}} $, and $ \mathit{TR}_{\mathrm{all}} $ denote the pooled numbers of TA, FA, FR, and TR across all repeated runs, respectively. Moreover, $ \mathit{FAR}_{\mathrm{all}} $ denotes the proportion of all invalid responses that are incorrectly accepted, $ \mathit{FRR}_{\mathrm{all}} $ denotes the proportion of all valid responses that are incorrectly rejected, and $ \mathit{Accuracy}_{\mathrm{all}} $ denotes the proportion of all evaluated responses that are classified correctly across all repeated runs. 

\Cref{tab:verifier_reliability} shows that Runs 1 and 4 achieved 100\% accuracy, while the remaining runs contained a small number of false decisions. Run 2 achieved 90.91\% accuracy, with a FAR of 0\% and an FRR of 14.29\%. Its only error was a FR in CoT-Step 5. The response correctly modeled hovering energy using the continuous dwell time at each assigned station. However, the Verifier additionally required an auxiliary variable to linearize the interaction between station assignment and dwell time. Since the step query did not require a linear or mixed-integer linear formulation, the original response was mathematically valid. Run 3 achieved 84.62\% accuracy, with a FAR of 14.29\% and an FRR of 16.67\%. Its false acceptance occurred in CoT-Step 4, where the Verifier accepted a formulation that accumulated computing and communication resource allocations over the whole mission horizon without ensuring that the required service had been completed before the declared completion slot. Its FR occurred in CoT-Step 5. The response already included flight, hovering, onboard-computing, uplink, and backhaul energy terms. However, the Verifier additionally required dwell-time linearization and more detailed unit declarations. Run 5 achieved 90.00\% accuracy, with a FAR of 0\% and an FRR of 14.29\%. Its FR occurred in CoT-Step 3, Attempt 1. The response used UAV-indexed execution-mode variables to identify a unique serving or relay UAV. It also specified that the same UAV handled the uplink and the subsequent onboard processing or backhaul forwarding. However, the Verifier required a separate first-hop indicator, although this auxiliary variable was not necessary for mathematical correctness. Run 6 achieved 93.33\% accuracy, with a FAR of 0\% and an FRR of 14.29\%. Its FR occurred in CoT-Step 1, Attempt 3. The response already included all required objective components. It also correctly represented deadline-based completion, finite flow time for unfinished tasks, and normalized computing and communication resource occupation. Nevertheless, the Verifier rejected it because the response did not explicitly state that every task arrived no later than the end of the mission horizon. It also used nonnegative rather than strictly positive coefficients for two reward terms and did not repeat the domain declaration for the UAV index in every UAV-indexed variable definition. These issues concerned notation and parameter-domain requirements rather than substantive errors in the objective. 

Overall, the Verifier evaluated 73 cases and produced 35 true acceptances, 33 true rejections, one false acceptance, and four FRs. The resulting FAR, FRR, and accuracy were 2.94\%, 10.26\%, and 93.15\%, respectively. These results indicate that the Verifier was generally reliable, particularly in preventing invalid formulations from being accepted. However, it was not fully reliable, as the higher FRR reveals a slightly conservative tendency to reject valid formulations. The FRs were mainly caused by additional linearization, notation, unit, or parameter-domain requirements that were not explicitly required by the corresponding step queries. The current Verifier uses GPT-5.6 Luna with low reasoning effort. Its reliability may be further improved by evaluating GPT-5.6 Terra or GPT-5.6 Sol, or by increasing the reasoning effort to medium or high. These configurations may support more careful judgments for complex mathematical formulations, but they may also increase inference cost and response latency. Therefore, future work should conduct a controlled comparison of model selection and reasoning-effort settings to determine whether the reliability improvement justifies the additional computational cost and latency. Future work will also investigate cross-model verification, in which the Responder and Verifier use different LLM families, such as ChatGPT and Claude. This approach may reduce correlated judgments and provide a more independent verification process.

\begin{table}[pos=htbp] 
    \centering
    \caption{Verifier reliability results over repeated runs.}
    \label{tab:verifier_reliability}
    \begingroup
    \setlength{\tabcolsep}{2.5pt}
    \renewcommand{\arraystretch}{1.08}

    \begin{tabularx}{\textwidth}{
        @{}
        >{\raggedright\arraybackslash}X
        >{\centering\arraybackslash}m{1.30cm}
        >{\centering\arraybackslash}m{1.3cm}   
        >{\centering\arraybackslash}m{1.3cm}    
        >{\centering\arraybackslash}m{1.3cm}   
        >{\centering\arraybackslash}m{1.3cm}
        >{\centering\arraybackslash}m{1.3cm}
        >{\centering\arraybackslash}m{1.3cm}
        >{\centering\arraybackslash}m{1.3cm}
        @{}
    }
        \toprule
        \textbf{Run}
        & \makecell{\textbf{Cases}}
        & \textbf{TA}
        & \textbf{FA}
        & \textbf{FR}
        & \textbf{TR}
        & \textbf{FAR}
        & \textbf{FRR}
        & \makecell{\textbf{Accuracy}} \\
        \midrule

        \textbf{Run 1}
        & \textbf{12}
        & \textbf{6}
        & \textbf{0}
        & \textbf{0}
        & \textbf{6}
        & \textbf{0\%}
        & \textbf{0\%}
        & \textbf{100\%} \\

        Run 2
        & 11 & 6 & 0 & 1 & 4 & 0\% & 14.29\% & 90.91\% \\

        Run 3
        & 13 & 5 & 1 & 1 & 6 & 14.29\% & 16.67\% & 84.62\% \\

        Run 4
        & 12 & 6 & 0 & 0 & 6 & 0\% & 0\% & 100\% \\

        Run 5
        & 10 & 6 & 0 & 1 & 3 & 0\% & 14.29\% & 90\% \\

        Run 6
        & 15 & 6 & 0 & 1 & 8 & 0\% & 14.29\% & 93.33\% \\

        \midrule

        \textbf{Overall}
        & \textbf{73} 
        & \textbf{35} 
        & \textbf{1} 
        & \textbf{4} 
        & \textbf{33} 
        & \textbf{2.94\%} 
        & \textbf{10.26\%} 
        & \textbf{93.15\%} \\ 

        \bottomrule
    \end{tabularx}
    \endgroup
\end{table}

\subsubsection{Quantitative Evaluation}

In \cref{tab:multiple_run_results}, the evaluations are defined as follows:
\paragraph{Formulation Correctness (FC)}
\begin{equation}
\label{eq:formulation_correctness}
 \mathit{FC} 
=
\frac{
\left| \mathit{Z}_{\mathrm{cor}}^{\mathrm{form}} \right|
}{
\left| \mathit{Z}_{\mathrm{req}}^{\mathrm{form}} \right|
}
\times 100\%.
\end{equation}
In \cref{eq:formulation_correctness}, $ \mathit{FC} $ denotes the formulation correctness, $ \mathit{Z}_{\mathrm{req}}^{\mathrm{form}} $ is the set of formulation groups required by the evaluation rubric, and $ \mathit{Z}_{\mathrm{cor}}^{\mathrm{form}} $ is the subset of required formulation groups that are generated with mathematically correct relations and the intended modeling semantics. The notation $|\cdot|$ denotes the cardinality of a set. Therefore, formulation correctness measures the percentage of all required formulation groups that are generated correctly.

\paragraph{Constraint Completeness/Recall (CC/CR)}
\begin{equation}
\label{eq:constraint_completeness}
 \mathit{CC} 
=
 \mathit{CR} 
=
\frac{
\left| \mathit{Z}_{\mathrm{cor}}^{\mathrm{con}} \right|
}{
\left| \mathit{Z}_{\mathrm{req}}^{\mathrm{con}} \right|
}
\times 100\%.
\end{equation}
In \cref{eq:constraint_completeness}, $ \mathit{CC} $ and $ \mathit{CR} $ denote constraint completeness and constraint recall, respectively, $ \mathit{Z}_{\mathrm{req}}^{\mathrm{con}} $ is the set of constraint groups required by the evaluation rubric, and $ \mathit{Z}_{\mathrm{cor}}^{\mathrm{con}} $ is the subset of required constraint groups that are correctly generated. Under the adopted rubric, constraint completeness and constraint recall are equivalent because both measure the percentage of required constraint groups that are correctly included in the generated formulation.

\paragraph{Constraint Precision (CP)}
\begin{equation}
\label{eq:constraint_precision}
 \mathit{CP} 
=
\frac{
\left| \mathit{Z}_{\mathrm{cor}}^{\mathrm{con}} \right|
}{
\left| \mathit{Z}_{\mathrm{gen}}^{\mathrm{con}} \right|
}
\times 100\%.
\end{equation}
In \cref{eq:constraint_precision}, $ \mathit{CP} $ denotes the constraint precision, $ \mathit{Z}_{\mathrm{gen}}^{\mathrm{con}} $ is the set of all constraint groups included in the generated formulation, and $ \mathit{Z}_{\mathrm{cor}}^{\mathrm{con}} $ is the subset of generated constraint groups that are mathematically correct, semantically valid, and within the scope of the target formulation. Therefore, constraint precision measures the percentage of generated constraint groups that are correct and relevant.

\paragraph{Hallucination Rate (HR)}
\begin{equation}
\label{eq:hallucination_rate}
 \mathit{HR} 
=
\frac{
\left| \mathit{Z}_{\mathrm{hal}}^{\mathrm{form}} \right|
}{
\left| \mathit{Z}_{\mathrm{gen}}^{\mathrm{form}} \right|
}
\times 100\%.
\end{equation}
In \cref{eq:hallucination_rate}, $ \mathit{HR} $ denotes the hallucination rate, $\mathit{Z}_{\mathrm{gen}}^{\mathrm{form}} $ is the set of all formulation groups included in the generated output, and $ \mathit{Z}_{\mathrm{hal}}^{\mathrm{form}} $ is the subset of generated formulation groups that contain unsupported content or substantive formulation errors. A formulation group is classified as hallucinated if it introduces an objective component, constraint, variable relationship, or physical assumption that is not justified by the specified problem requirements, or if it contains a substantive mathematical or logical error that conflicts with the intended system requirements. A substantive error is one that changes the mathematical meaning, feasible region, physical relationship, or constraint logic of the formulation. Minor notation, wording, or formatting issues that do not change the intended mathematical semantics are not counted as hallucinations. Each formulation group is counted at most once in $ \mathit{Z}_{\mathrm{hal}}^{\mathrm{form}} $. Therefore, the hallucination rate measures the percentage of generated formulation groups that contain unsupported content or substantive formulation errors.

\paragraph{Semantic Consistency with Retrieved RAG Evidence}
\begin{equation}
\label{eq:rag_semantic_consistency}
 \mathit{SC}_{\mathrm{RAG}} 
=
\frac{
\left| \mathit{Z}_{\mathrm{sup}}^{\mathrm{RAG}} \right|
}{
\left| \mathit{Z}_{\mathrm{evd}}^{\mathrm{form}} \right|
}
\times 100\%.
\end{equation}
In \cref{eq:rag_semantic_consistency}, $ \mathit{SC}_{\mathrm{RAG}} $ denotes the semantic consistency with retrieved RAG evidence, $ \mathit{Z}_{\mathrm{evd}}^{\mathrm{form}} $ is the set of generated formulation groups that require evidence support, and $ \mathit{Z}_{\mathrm{sup}}^{\mathrm{RAG}} $ is the subset whose core mathematical and physical semantics are supported by the retrieved non-target evidence from the RAG database. Semantic support does not require identical notation or equation formatting; it requires consistency in the decision meaning, modeling function, physical relationship, and constraint logic. Therefore, this metric measures the percentage of evidence-requiring formulation groups that are semantically supported by the retrieved RAG evidence.

\paragraph{Formulation Groups for Evaluation}

For evaluation, the target formulation is organized into 11 functionally coherent formulation groups. One group represents the optimization objective and its associated normalized resource-occupation quantities in \cref{eq:obj_revised}. The remaining ten constraint groups represent: 
1) station assignment and collection-credit linkage in \cref{eq:C1_aligned}; 
2) route-flow and depot-balance constraints in \cref{eq:C3_aligned,eq:C4_aligned};
3) payload, travel-distance, mission-timing, and service-duration constraints in \cref{eq:C6C7_aligned,eq:depot_dep_horizon_merged,eq:travel_time_merged1,eq:travel_time_merged2,eq:travel_time_merged3,eq:C17_aligned}; 
4) execution-mode selection and associated-UAV selection for UAV-assisted modes in \cref{eq:mode_select_aligned}; 
5) completion-status and completion-time encoding and deadline-feasibility constraints in \cref{eq:Tcomp_def1,eq:Tcomp_def2,eq:deadline_indicator}, with the resulting completion-time quantity used in the flow-time term of \cref{eq:obj_revised}; 
6) computing-capacity constraints in \cref{eq:cap_compute}; 
7) communication-capacity constraints and mode/path-consistency constraints for computing and communication allocations in \cref{eq:cap_comm,eq:mode_gate_comp,eq:mode_gate_comm}; 
8) completion-dependent cumulative service-sufficiency constraints in \cref{eqn:suff,eqn:suff4}; 
9) service-window, UAV-presence, and UAV-involved processing constraints in \cref{eq:eta_one_station_aligned,eq:eta_assign_aligned,eq:eta_time_merged_indicator_1,eq:eta_time_merged_indicator_2,eq:sw_ul_gate_aligned}; and 
10) UAV energy accounting and energy-budget constraints in \cref{eqn_energy1,eqn_energy2,eqn_energy3,eqn_energy4}. A formulation group is defined as a set of equations that jointly implements one distinct modeling function. A constraint group is a formulation group whose equations jointly define one feasibility requirement or one family of closely related feasibility requirements. We evaluate equation groups rather than individual equations because the same modeling function can be represented by a single compact equation, several mode-specific equations, or an algebraically equivalent formulation with auxiliary variables. Counting individual equations would therefore over-weight modeling functions expressed through many equations and under-weight compact but semantically equivalent formulations. Group-level evaluation instead measures whether each required modeling function is present, mathematically correct, semantically supported, and consistent with the target system requirements.

\Cref{tab:multiple_run_results} summarizes the formulation correctness (FC), constraint completeness (CC), constraint precision (CP), hallucination rate (HR), semantic consistency with the retrieved RAG evidence ($\mathit{SC}_{\mathrm{RAG}} $), reported cost, latency, and total token usage over six repeated runs of the proposed RAG + CoT + Verifier framework. Runs 1, 2, 4, 5, and 6 achieved 100\% FC, CC, and CP. They also achieved an HR of 0\% and an $ \mathit{SC}_{\mathrm{RAG}} $ of 100\%. Therefore, five of the six runs correctly generated all 11 required formulation groups and all 10 required constraint groups. These runs contained no unsupported or substantively incorrect formulation group. Run 3 was the only exception. It achieved an FC of 90.91\%, because 10 of the 11 required formulation groups were correctly generated. Its CC and CP were both 90.00\%, because 9 of the 10 required constraint groups were correctly formulated. The completion-dependent cumulative service-sufficiency group contained a substantive temporal-consistency error. Specifically, the generated constraints did not ensure that the required computing and communication services had been completed by the declared task-completion slot. Therefore, this group was classified as a substantively incorrect formulation group, resulting in an HR of 9.09\%. The same group was not semantically consistent with the retrieved RAG evidence, resulting in an $ \mathit{SC}_{\mathrm{RAG}} $ of 90.91\%. Across the six runs, the reported cost ranged from USD~0.69 to USD~1.33. The latency ranged from 241.99~s to 368.97~s, while the total token usage ranged from 2,232,071 to 5,001,774 tokens. These differences were partly associated with the internal interactions between the Responder and the Verifier. As shown in \cref{tab:verifier_reliability}, the number of evaluated cases ranged from 10 to 15. Each case represents one candidate response evaluated by the Verifier. Therefore, a larger number of cases generally indicates that more candidate responses were evaluated during the internal responder-verifier interaction, often because a rejected response triggered a revision by the Responder. For example, Run 5 had the fewest cases and also had the lowest cost, latency, and token usage, whereas Run 6 had the most cases and incurred the highest cost and token usage. Thus, the case count provides an approximate indication of the interaction and verification demand in each run, although it is not a direct count of the correction iterations. 

Overall, the repeated-run results demonstrate the effectiveness and repeatability of the proposed RAG + CoT + Verifier framework for the evaluated formulation task. High formulation quality was achieved using GPT-5.6 Luna with low reasoning effort. In principle, more capable models, such as GPT-5.6 Terra or GPT-5.6 Sol, or medium and high reasoning-effort settings, may improve the evaluation of difficult mathematical formulations and reduce occasional formulation or verification errors. However, these configurations may also increase cost, latency, and token usage. Since they were not evaluated in the current experiments, their performance improvements cannot yet be confirmed. Future work should conduct controlled comparisons to evaluate the trade-off among formulation quality, Verifier reliability, cost, latency, and token consumption. Future work can also extend the current evaluation with retrieval-level metrics, such as context relevance and retrieval recall, and generation-level metrics, such as faithfulness, correctness, and answer relevance \cite{ref_quant_evalu1,ref_quant_evalu2}. It can further examine sentence- or span-level hallucination detection \cite{ref_quant_evalu3,ref_quant_evalu4}, inter-rater agreement among multiple human evaluators \cite{ref_quant_evalu1}, and robustness across different prompts, tasks, external knowledge sources, and model families \cite{ref_quant_evalu3,ref_quant_evalu4}. These evaluations would provide a broader assessment of retrieval quality, factual grounding, evaluation consistency, and generalizability.

\begin{table}[pos=htbp]
    \centering
    \caption{Multiple-run evaluation results for the proposed RAG + CoT + Verifier framework.}
    \label{tab:multiple_run_results}
    \begingroup
    \setlength{\tabcolsep}{0.8pt}
    \renewcommand{\arraystretch}{1.15}
    \renewcommand{\tabularxcolumn}[1]{m{#1}}

    \begin{tabularx}{\textwidth}{
        @{}
        >{\raggedright\arraybackslash}m{1.25cm}
        *{8}{>{\centering\arraybackslash}X}
        @{}
    }
        \toprule

        \makecell[l]{\textbf{Run}}
        & \makecell[c]{\textbf{FC}}
        & \makecell[c]{\textbf{CC}}
        & \makecell[c]{\textbf{CP}}
        & \makecell[c]{\textbf{HR}}
        & \makecell[c]{\textbf{$ \mathit{SC}_{\mathrm{RAG}} $}}
        & \makecell[c]{\textbf{Cost (USD)}}
        & \makecell[c]{\textbf{Latency (s)}}
        & \makecell[c]{\textbf{Token Used}} \\

        \midrule

        \textbf{Run 1}
        & \textbf{100\%}    
        & \textbf{100\%}    
        & \textbf{100\%}    
        & \textbf{0\%}      
        & \textbf{100\%}    
        & \textbf{0.99}      
        & \textbf{368.97}      
        & \textbf{3,720,094} \\   

        Run 2
        & 100\%    
        & 100\%    
        & 100\%    
        & 0\%      
        & 100\%    
        & 0.92      
        & 279.38      
        & 3,201,760 \\   

        Run 3
        & 90.91\%       
        & 90\%       
        & 90\%       
        & 9.09\%       
        & 90.91\%       
        & 1.01       
        & 321.82       
        & 3,601,905 \\    

        Run 4
        & 100\%       
        & 100\%       
        & 100\%       
        & 0\%       
        & 100\%       
        & 0.98       
        & 306.14       
        & 3,509,420 \\    

        Run 5
        & 100\%       
        & 100\%       
        & 100\%       
        & 0\%       
        & 100\%       
        & 0.69       
        & 241.99       
        & 2,232,071  \\    

        Run 6
        & 100\%       
        & 100\%       
        & 100\%       
        & 0\%       
        & 100\%       
        & 1.33       
        & 368.70       
        &  5,001,774  \\    
    
        \bottomrule
    \end{tabularx}
    \endgroup
\end{table}

\subsubsection{Ablation Study}

\begin{table}[pos=htbp]
    \centering
    \caption{Formulation quality and ablation results}
    \label{tab:formulation_quality_ablation}
    \begingroup
    \setlength{\tabcolsep}{1.5pt}
    \renewcommand{\arraystretch}{1.12}
    \renewcommand{\tabularxcolumn}[1]{m{#1}}

    \begin{tabularx}{\textwidth}{
        @{}
        >{\raggedright\arraybackslash}m{3.70cm}
        >{\centering\arraybackslash}m{2.45cm}
        >{\centering\arraybackslash}m{2.35cm}
        >{\centering\arraybackslash}m{2.20cm}
        >{\centering\arraybackslash}m{2.00cm}
        >{\centering\arraybackslash}m{3.00cm}
        @{}
    }
        \toprule

        \textbf{Method}

        & \makecell[c]{\textbf{FC}} 
        & \makecell[c]{\textbf{CC}}
        & \makecell[c]{\textbf{CP}}
        & \makecell[c]{\textbf{HR}}
        & \makecell[c]{\textbf{$ \mathit{SC}_{\mathrm{RAG}} $}} \\

        \midrule

        Direct LLM
        & 54.55\%        
        & 60\%           
        & 60\%           
        & 45.45\%         
        & N/A \\        

        RAG-only
        & 72.73\%        
        & 80\%           
        & 80\%           
        & 27.27\%         
        & 81.82\% \\     

        RAG + CoT
        & 90.91\%        
        & 90\%           
        & 90\%           
        & 9.09\%         
        & 90.91\% \\     
        \bottomrule
    \end{tabularx}
    \endgroup
\end{table}

In the ablation study, we progressively removed individual components from the proposed RAG + CoT + Verifier framework. First, we removed the Verifier while retaining RAG and CoT, resulting in the RAG + CoT variant. Next, we removed both the Verifier and CoT, leaving only the retrieval component; this variant is referred to as RAG-only. Finally, we removed the Verifier, CoT, and RAG, allowing the LLM to generate the formulation directly from the user prompt; this baseline is referred to as Direct LLM. As shown in \cref{tab:formulation_quality_ablation}, formulation quality improved progressively from Direct LLM to RAG-only and then to RAG + CoT. The Direct LLM baseline correctly generated only 6 of the 11 equation groups, yielding an FC of 54.55\%, a CC and CP of 60.00\%, and an HR of 45.45\%. For example, its cumulative-service constraints evaluated computing and communication allocations over the entire mission horizon rather than up to the declared completion slot. This allowed a task to be marked as completed before its required service was delivered. It also omitted the separate binary deadline-miss penalty and introduced an unsupported service-window parameter. Because this mode made no retrieval calls, its $ \mathit{SC}_{\mathrm{RAG}} $ is reported as N/A. Adding RAG increased FC to 72.73\%, CC and CP to 80.00\%, and $ \mathit{SC}_{\mathrm{RAG}} $ to 81.82\%, while reducing HR to 27.27\%. However, RAG-only retained the cumulative-service timing error and did not properly link hovering energy to station assignment. Its $ \mathit{SC}_{\mathrm{RAG}} $ exceeded its FC because the retrieved knowledge supported separate computing and communication resource weights as a general modeling variant, although this design did not match the target formulation's combined resource-occupation weight. RAG + CoT achieved the best performance, with an FC and $ \mathit{SC}_{\mathrm{RAG}} $ of 90.91\%, a CC and CP of 90.00\%, and an HR of 9.09\%. Its only substantive error occurred in the mission-timing constraints, where an unsupported upper bound was added to the required lower-bound arrival-time relation. This forced equality between consecutive travel times and incorrectly excluded feasible solutions in which a UAV waits before arriving at the next station. Overall, RAG improved domain coverage, while stepwise CoT further reduced unsupported components and cross-equation consistency errors.

\subsubsection{Comparison of the User Prompts} 

To examine how user-prompt decomposition affects the proposed RAG + CoT + Verifier framework, we designed 3 user prompts with different numbers of CoT steps, namely user-prompt-1, user-prompt-2, and user-prompt-3. The considered optimization problem contains an objective function, a UAV Routing Model, an MEC Service Model, and a notation summary. The proposed user-prompt-1 contains six CoT steps: defining the objective function; deriving the UAV-routing constraints; deriving Part 1, Part 2, and Part 3 of the MEC-service constraints in three separate steps; and summarizing the mathematical notation and symbol definitions. This design divides the relatively long MEC Service Model into smaller tasks. Therefore, the Responder does not need to generate an excessively long answer in one query, and the Verifier can identify errors within a more focused set of constraints. User-prompt-2 contains four CoT steps: defining the objective function, deriving the UAV Routing Model, deriving the complete MEC Service Model, and summarizing the notation. This model-level decomposition preserves a clear boundary between routing and MEC scheduling. It also allows the two models to be formulated and verified independently while using fewer interactions than User Prompt 1. User-prompt-3 contains three CoT steps: defining the objective function, jointly deriving the UAV Routing Model and the MEC Service Model, and summarizing the notation. Its advantage is that the Responder can consider the coupling between UAV routing and MEC scheduling within the same query, which reduces cross-step references and helps maintain the routing-induced service-window relationships. 

They all can achieve an FC of \(100\%\), a CC of \(100\%\), a CP of \(100\%\), an HR of \(0\%\), and an \( \mathit{SC}_{\mathrm{RAG}} \) of \(100\%\). However, the Verifier reliability tests for user-prompt-2 and user-prompt-3 showed that the Verifier accuracy for these broader prompt designs could fall below \(84.62\%\). A coarser prompt decomposition combines more modeling requirements and constraint groups into each query, which increases the complexity of the generated response and the evaluation burden on the Verifier. Consequently, the Verifier may overlook a localized error or incorrectly reject an otherwise valid formulation. Broader queries also provide fewer intermediate verification checkpoints and make local errors more difficult to isolate. Therefore, user-prompt-1 is proposed in this paper.

\subsection{Effectiveness of the Hierarchical DRL Approach}

To clarify the scope of this evaluation, the following experiments are designed as a controlled assessment of the proposed sequential hierarchy rather than an exhaustive comparison of all reinforcement-learning architectures. The upper-layer PPO policy is trained first and then fixed. The lower-layer PPO and A2C methods use the same routing output, service windows, state representation, factorized action representation, feasibility masks, and training environment. This setting holds the routing-induced inputs, environment model, lower-layer state and action representations, and feasibility rules constant, while PPO and A2C use their respective policy-update mechanisms and algorithm-specific hyperparameters. PPO and A2C are selected because they can share the same masked policy representation in the adopted implementation. In contrast, the lower-layer action contains discrete execution-mode selection together with computation and communication resource allocation under state-dependent feasibility masks. Standard soft actor-critic (SAC) and twin delayed deep deterministic policy gradient (TD3) implementations are therefore not direct replacements for the adopted policy structure \cite{Ref_DRL_state1,Ref_DRL_state2}. A fair comparison would require a specifically designed hybrid actor-critic structure, a compatible replay mechanism, and mask-aware target computation. Similarly, a non-hierarchical PPO would require a unified multi-timescale policy that jointly represents routing, task execution modes, and resource allocations. This would produce a substantially larger joint action space and require a separate state, reward, mask, and training design \cite{Ref_DRL_state3}. Accordingly, the following PPO-A2C comparison is designed to characterize the training behavior and solution feasibility of the two lower-layer learning methods under the same hierarchical configuration. This controlled comparison provides a focused basis for examining the behavior of the two lower-layer learning methods under the proposed hierarchical setting. A broader evaluation can further extend this controlled assessment and provide additional insight into the proposed hierarchical framework. The experiments are conducted under a controlled sequential PPO-based hierarchy, where the upper-layer routing policy provides fixed routing-induced service conditions for lower-layer scheduling. Building on this evaluation, future work will investigate a wider range of routing and scheduling strategies. These will include heuristic upper-layer routing, a greedy lower-layer scheduler, monolithic or non-hierarchical PPO, rolling-horizon optimization-based offloading, and mask-aware hybrid SAC and TD3 baselines. Future evaluations will also consider independent multi-seed experiments with significance tests and mean-and-standard-deviation reporting, as well as out-of-distribution task arrivals and different system sizes. 
According to the evaluation scope above, the effectiveness of the upper-layer and lower-layer DRL approach are discussed as follows. 

\subsubsection{Upper-layer DRL Training Results} 
\begin{figure}[pos=htbp]
  \centering
  \includegraphics[width=0.5\columnwidth]{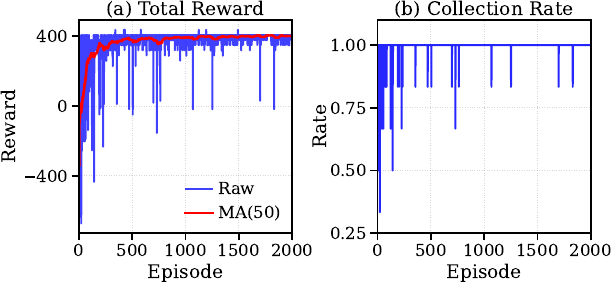}
  \caption{Upper-layer DRL training results with PPO. The results include the total reward and the collection rate.}
  \label{fig:upper_reward_collection}
\end{figure}

In \cref{fig:upper_reward_collection} (a), the upper-layer PPO shows good learning ability and convergence. \textit{Raw} denotes the reward of each episode and \textit{MA(50)} is the moving average of the reward of the last 50 episodes, which can better reflect the overall training trend. In the early stage, the \textit{Raw} reward fluctuates sharply, indicating active exploration. With training, \textit{MA(50)} increases rapidly and then stabilizes around 400, indicating that the policy is improved rapidly and converges to a stable high reward solution. Although occasional drops still appear in the \textit{Raw} reward, the stable \textit{MA(50)} suggests that the overall training performance is robust. \Cref{fig:upper_reward_collection} (b) further shows that the learned upper-layer policy achieves a high collection rate. After several early exploratory fluctuations, the collection rate quickly approaches $100\%$ and remains at or near $100\%$ in most episodes, indicating that the PPO policy can reliably plan UAV routing and complete almost all collection tasks. Although a few occasional drops still appear, the overall result remains highly stable. 

\subsubsection{Lower-layer DRL Training Results}
\begin{figure}[pos=htbp]
  \centering
  \includegraphics[width=0.5\columnwidth]{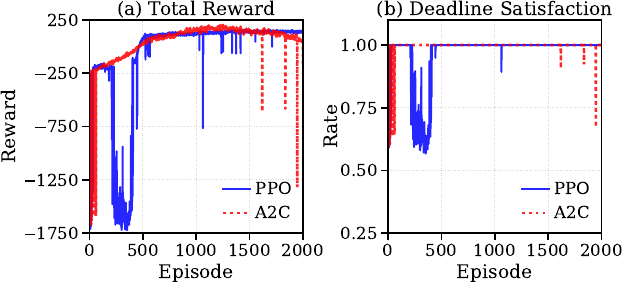}
  \caption{Lower-layer DRL training results with PPO and A2C. The results include the total reward and the deadline satisfaction rate.}
  \label{fig:lower_reward_deadline}
\end{figure}

\Cref{fig:lower_reward_deadline} compares the lower-layer DRL training performance of PPO and A2C in terms of total reward and deadline satisfaction rate, where the deadline satisfaction rate denotes the fraction of tasks completed within their deadlines in each episode. \Cref{fig:lower_reward_deadline} (a) shows that both PPO and A2C improve the lower-layer policy from highly negative rewards to a converged positive-reward region. PPO exhibits larger fluctuations in the intermediate stage, but its reward quickly recovers and remains relatively stable afterward. In contrast, the total reward under A2C increases more smoothly in the early stage, but several sharp reward drops still appear in the later stage, indicating weaker stability after convergence. \Cref{fig:lower_reward_deadline} (b) shows that both methods achieve a high deadline satisfaction rate after training. However, PPO maintains a rate very close to $100\%$ more consistently in the later stage, whereas A2C still experiences several late-stage drops and cannot always sustain the ideal $100\%$ deadline satisfaction rate. Based on these results, PPO is preferred for the lower-layer DRL training. The main reason is that PPO provides more robust converged performance, with more stable rewards and more reliable deadline satisfaction in the later training stage. In particular, unlike A2C, PPO is able to maintain nearly perfect deadline satisfaction in most converged episodes. These results are obtained under the service windows generated by the fixed upper-layer routing policy. Therefore, they demonstrate the effectiveness and stability of lower-layer PPO under the adopted sequential hierarchy. However, this experiment does not evaluate whether alternative upper-layer routes could further improve the coupled routing-MEC performance, because the purpose here is to isolate the lower-layer scheduling performance under the fixed service windows generated by the trained upper-layer policy. Evaluating alternative upper-layer routes would require an additional feedback-aware route-refinement procedure, which is beyond the adopted sequential hierarchy and is left for future work.

\section{Conclusion}
\label{sec:Conclusion}

In this paper, we have studied a hybrid coordination problem in CMfg, where UAV-assisted product collection is tightly coupled with MEC task processing for industrial sensor devices. To address the difficulty of formulating such a problem, we have proposed an interactive agentic AI framework that integrates LLMs, RAG, and CoT reasoning to support interpretable mathematical modeling. To solve the resulting optimization problem, we have further developed a hierarchical DRL approach based on PPO, in which the upper layer handles UAV routing and the lower layer performs computational task scheduling and resource allocation under coupled operational constraints. Simulation results in the evaluated scenarios demonstrate the effectiveness of the proposed framework under the sequential hierarchy. The proposed design provides a tractable way to combine agentic-AI-assisted formulation with hierarchical PPO-based optimization. In particular, the upper layer generates feasible UAV routes, service windows, and residual-energy information, while the lower layer uses these routing-induced inputs to perform MEC task scheduling and resource allocation. The reported routing and deadline-satisfaction results illustrate the behavior and feasibility of the framework within the evaluated simulation scenario. In the sequential implementation, the information flow is mainly from the routing layer to the MEC scheduling layer, and lower-layer MEC indicators are not yet used to refine the upper-layer routing policy. A promising future direction is to develop feedback-aware or iterative hierarchical training, where deadline satisfaction, backlog reduction, and MEC reward can guide upper-layer route refinement while maintaining training stability. Future studies should also evaluate resource-constrained and larger-scale systems, heterogeneous UAV fleets, more realistic wireless channel models, and more dynamic task-arrival processes. In particular, future resource-stress experiments should examine the scheduling trade-offs caused by limited computing and communication capacities, increased task demand, and tighter deadline requirements. The ablation analysis quantifies the effects of RAG, CoT reasoning, and the Verifier under the evaluated setting. Future studies should extend this analysis to alternative verification designs and hierarchical-decomposition baselines. The evaluation is conducted under one representative problem configuration. Therefore, the reported results should not be interpreted as empirical evidence of generalization across different system sizes or constraint sets. Future studies should systematically evaluate the hierarchical-DRL solution performance under varying numbers of UAVs and manufacturing stations, as well as different combinations of routing, energy, communication, computation, service-window, and deadline constraints. They should also consider heterogeneous UAV fleets, more realistic wireless channel models, and more dynamic task-arrival processes.

\section*{CRediT authorship contribution statement}
\textbf{Hanwen Zhang}: Conceptualization, Formal analysis, Investigation, Methodology, Software, Validation, Visualization, Writing -- original draft, Writing -- review \& editing.
\textbf{Dusit Niyato}: Conceptualization, Methodology, Funding acquisition, Project administration, Resources, Supervision, Writing -- review \& editing. 
\textbf{Wei Zhang}: Funding acquisition, Project administration, Resources, Supervision, Writing -- review \& editing. 
\textbf{Xin Lou}: Writing -- review \& editing. 
\textbf{Malcolm Yoke Hean Low}: Writing -- review \& editing. 

\section*{Data availability}
Data is available in \cref{sec:Dataset_Construction}.

\section*{Declaration of competing interest}

The authors declare that they have no known competing financial interests or personal relationships that could have appeared to influence the work reported in this paper.

\section*{Acknowledgments}
This work was supported in part by the Agency for Science, Technology and Research (A*STAR) under its Manufacturing, Trade \& Connectivity (MTC) Individual Research Grants (IRG) under Grant M23M6c0113; in part by MTC Programmatic under Grant M23L9b0052; in part by the Seatrium New Energy Laboratory, Singapore Ministry of Education (MOE) Tier 1, under Grant RT5/23 and Grant RG24/24; in part by the Nanyang Technological University (NTU) Center for Computational Technologies in Finance (NTU-CCTF); and in part by the Research Innovation and Enterprise (RIE) 2025 Industry Alignment Fund-Industry Collaboration Projects (IAF-ICP), administered by A*STAR, under Grant I2301E0026. (\textit{Corresponding author: Wei Zhang.)}

\appendix
\section{Appendix}
\label{sec:Appendix}
\subsection{Dataset Construction}
\label{sec:Dataset_Construction}

To validate the proposed framework, we construct a domain-specific RAG knowledge base consisting of 11 Markdown files. The knowledge base organizes reusable modeling principles and formulation primitives related to system modeling, objective design, UAV routing, MEC task offloading, computing and communication resources, task completion and deadlines, service-window coupling, UAV energy modeling, and system-boundary variants. In the implementation, these Markdown files are stored in the \texttt{database/} directory and indexed for semantic retrieval by the Responder and Verifier. The reference formulation, including the objective in \cref{eq:obj_revised} and the constraint groups in 
\crefrange{eq:C1_aligned}{eq:C4_aligned} and \crefrange{eq:C6C7_aligned}{eqn_energy4}, is maintained separately for evaluating the generated formulations. Our code and dataset for the proposed agentic AI framework are uploaded to GitHub\footnote{\url{https://github.com/Puppet88/Agentic-AI-UAV}} and released under the MIT License. The repository includes a \texttt{requirements.txt} file listing the required Python packages and a README file with reproduction instructions. The README documents the software environment, required local files, API-key setup, and execution procedure.


\printcredits

\bibliographystyle{elsarticle-num}

\bibliography{Refbib.bib}

@article{Ref_Elsevier_IoT1,
title = {RAAF-MEC: Reliable and anonymous authentication framework for IoT-enabled mobile edge computing environment},
journal = {Internet of Things},
volume = {29},
pages = {101459},
year = {2025},
issn = {2542-6605},
doi = {https://doi.org/10.1016/j.iot.2024.101459},
url = {https://www.sciencedirect.com/science/article/pii/S2542660524004001},
author = {Omar Alruwaili and Muhammad Tanveer and Saud Alhajaj Aldossari and Saad Alanazi and Ammar Armghan}
}

@article{Ref_Elsevier_IoT2,
title = {CORAL: cognitive optimization for robust and fair latency in non-ideal IRS-assisted UAV-MEC IoT networks},
journal = {Internet of Things},
volume = {38},
pages = {101971},
year = {2026},
issn = {2542-6605},
doi = {https://doi.org/10.1016/j.iot.2026.101971},
url = {https://www.sciencedirect.com/science/article/pii/S2542660526001010},
author = {Mofadal Alymani and Mohammed Aljebreen and Manal {Abdullah Alohali} and Mohammed A. AlAqil and Nadhem Nemri and Rakan Alanazi and Nouf {Atiahallah Alghanmi}}
}

@article{Ref_Elsevier_IoT3,
title = {Digital twin-driven semantic offloading for LEO-MEC-enabled remote IoT networks},
journal = {Internet of Things},
volume = {35},
pages = {101831},
year = {2026},
issn = {2542-6605},
doi = {https://doi.org/10.1016/j.iot.2025.101831},
url = {https://www.sciencedirect.com/science/article/pii/S2542660525003452},
author = {Ayalneh {Bitew Wondmagegn} and Dongwook Won and Donghyun Lee and Jaemin Kim and Juyoung Kim and Sungrae Cho}
}

@article{Ref_Elsevier_IoT4,
title = {Reliability-oriented dynamic task offloading for time-varying satellite IoT: A lightweight multi-agent cooperative strategy optimization method},
journal = {Internet of Things},
volume = {32},
pages = {101603},
year = {2025},
issn = {2542-6605},
doi = {https://doi.org/10.1016/j.iot.2025.101603},
url = {https://www.sciencedirect.com/science/article/pii/S2542660525001167},
author = {Xin-tong Pei and Zhen-jiang Zhang and Qing-an Zeng and Ying-si Zhao}
}

@ARTICLE{Ruichen2,
  author={Zhang, Ruichen and Du, Hongyang and Liu, Yinqiu and Niyato, Dusit and Kang, Jiawen and Sun, Sumei and Shen, Xuemin and Poor, H. Vincent},
  journal={IEEE Network}, 
  title={Interactive AI With Retrieval-Augmented Generation for Next Generation Networking}, 
  year={2024},
  volume={38},
  number={6},
  pages={414-424},
  doi={10.1109/MNET.2024.3401159}}

@ARTICLE{JSP_DRL18_CMfg,
  author={Wang, Xiaohan and Laili, Yuanjun and Zhang, Lin and Liu, Yongkui},
  journal={IEEE Transactions on Automation Science and Engineering}, 
  title={Hybrid Task Scheduling in Cloud Manufacturing With Sparse-Reward Deep Reinforcement Learning}, 
  year={2025},
  volume={22},
  number={},
  pages={1878-1892},
  doi={10.1109/TASE.2024.3371250}
}

@ARTICLE{JSP_DRL17,
  author={Li, Yuxin and Wang, Qingzheng and Li, Xinyu and Gao, Liang and Fu, Ling and Yu, Yanbin and Zhou, Wei},
  journal={IEEE Transactions on Systems, Man, and Cybernetics: Systems}, 
  title={Real-Time Scheduling for Flexible Job Shop With AGVs Using Multiagent Reinforcement Learning and Efficient Action Decoding}, 
  year={2025},
  volume={55},
  number={3},
  pages={2120-2132},
  doi={10.1109/TSMC.2024.3520381}}

@article{JSP_DRL18_R1,
title = "Cloud manufacturing: A new service-oriented networked manufacturing model",
author = "Li, \{Bo Hu\} and Lin Zhang and Wang, \{Shi Long\} and Fei Tao and Cao, \{Jun Wei\} and Jiang, \{Xiao Dan\} and Xiao Song and Chai, \{Xu Dong\}",
year = "2010",
month = jan,
language = "English",
volume = "16",
pages = "1--7+16",
journal = "Jisuanji Jicheng Zhizao Xitong/Computer Integrated Manufacturing Systems, CIMS",
issn = "1006-5911",
publisher = "CIMS",
number = "1",
url = {https://research.buaa.edu.cn/zh/publications/cloud-manufacturing-a-new-service-oriented-networked-manufacturin/}
}

@article{UseCase2_R3,
   author = {Satoglu, Sule Itir and Sahin, I. Ethem},
   title = {Design of a just-in-time periodic material supply system for the assembly lines and an application in electronics industry},
   journal = {The International Journal of Advanced Manufacturing Technology},
   volume = {65},
   number = {1},
   pages = {319-332},
   ISSN = {1433-3015},
   DOI = {10.1007/s00170-012-4171-7},
   url = {https://doi.org/10.1007/s00170-012-4171-7},
   year = {2013},
   type = {Journal Article}
}

@article{UseCase2_R13,
   author = {Mohsan, Syed Agha Hassnain and Othman, Nawaf Qasem Hamood and Li, Yanlong and Alsharif, Mohammed H. and Khan, Muhammad Asghar},
   title = {Unmanned aerial vehicles (UAVs): practical aspects, applications, open challenges, security issues, and future trends},
   journal = {Intelligent Service Robotics},
   volume = {16},
   number = {1},
   pages = {109-137},
   ISSN = {1861-2784},
   DOI = {10.1007/s11370-022-00452-4},
   url = {https://doi.org/10.1007/s11370-022-00452-4},
   year = {2023},
   type = {Journal Article}
}

@article{UseCase2_R13b,
title = {Embracing drones and the Internet of drones systems in manufacturing – An exploration of obstacles},
journal = {Technology in Society},
volume = {78},
pages = {102648},
year = {2024},
issn = {0160-791X},
doi = {https://doi.org/10.1016/j.techsoc.2024.102648},
url = {https://www.sciencedirect.com/science/article/pii/S0160791X24001969},
author = {Dauren Askerbekov and Jose Arturo Garza-Reyes and Ranjit {Roy Ghatak} and Rohit Joshi and Jayakrishna Kandasamy and Daniel {Luiz de Mattos Nascimento}}
}

@ARTICLE{TwoLayerDRL,
  author={Mao, Xiao and Wu, Guohua and Fan, Mingfeng and Cao, Zhiguang and Pedrycz, Witold},
  journal={IEEE Transactions on Automation Science and Engineering}, 
  title={DL-DRL: A Double-Level Deep Reinforcement Learning Approach for Large-Scale Task Scheduling of Multi-UAV}, 
  year={2025},
  volume={22},
  number={},
  pages={1028-1044},
  doi={10.1109/TASE.2024.3358894}}

@article{priority_scores1,
title = {Reinforcement learning with priority decentralized PPO for multi-vessel cooperative rescue scheduling in flood disaster},
journal = {Alexandria Engineering Journal},
volume = {138},
pages = {96-113},
year = {2026},
issn = {1110-0168},
doi = {https://doi.org/10.1016/j.aej.2026.01.047},
url = {https://www.sciencedirect.com/science/article/pii/S1110016826000761},
author = {Yufeng Zhou and Wei Yang and Ying Gong}
}

@ARTICLE{priority_scores2,
  author={Lee, Hyunsung and Lee, Jinkyu and Yeom, Ikjun and Woo, Honguk},
  journal={IEEE Access}, 
  title={Panda: Reinforcement Learning-Based Priority Assignment for Multi-Processor Real-Time Scheduling}, 
  year={2020},
  volume={8},
  number={},
  pages={185570-185583},
  doi={10.1109/ACCESS.2020.3029040}}

@ARTICLE{Routing_action1,
  author={Fan, Mingfeng and Wu, Yaoxin and Liao, Tianjun and Cao, Zhiguang and Guo, Hongliang and Sartoretti, Guillaume and Wu, Guohua},
  journal={IEEE Transactions on Vehicular Technology}, 
  title={Deep Reinforcement Learning for UAV Routing in the Presence of Multiple Charging Stations}, 
  year={2023},
  volume={72},
  number={5},
  pages={5732-5746},
  doi={10.1109/TVT.2022.3232607}}

@article{Ref_Related_UAV1,
title = {Routing a fleet of unmanned aerial vehicles: A trajectory optimisation-based framework},
journal = {Transportation Research Part B: Methodological},
volume = {200},
pages = {103312},
year = {2025},
issn = {0191-2615},
doi = {https://doi.org/10.1016/j.trb.2025.103312},
url = {https://www.sciencedirect.com/science/article/pii/S0191261525001614},
author = {Walton P. Coutinho and Jörg Fliege and Maria Battarra and Anand Subramanian}
}

@article{Ref_Related_UAV2,
title = {A dynamic drone routing problem with uncertain demand and energy consumption},
journal = {Transportation Research Part B: Methodological},
volume = {202},
pages = {103335},
year = {2025},
issn = {0191-2615},
doi = {https://doi.org/10.1016/j.trb.2025.103335},
url = {https://www.sciencedirect.com/science/article/pii/S0191261525001845},
author = {Guilherme O. Chagas and Leandro C. Coelho and Demetrio Laganà and Patrizia Beraldi}
}

@article{Ref_Related_UAV3,
title = {C-SPPO: A deep reinforcement learning framework for large-scale dynamic logistics UAV routing problem},
journal = {Chinese Journal of Aeronautics},
volume = {38},
number = {5},
pages = {103229},
year = {2025},
issn = {1000-9361},
doi = {https://doi.org/10.1016/j.cja.2024.09.005},
url = {https://www.sciencedirect.com/science/article/pii/S1000936124003662},
author = {Fei WANG and Honghai ZHANG and Sen DU and Mingzhuang HUA and Gang ZHONG}
}

@article{Ref_Related_UAV4,
title = {Collaborative vessel–unmanned aerial vehicle routing for time-window-constrained offshore parcel delivery},
journal = {Transportation Research Part C: Emerging Technologies},
volume = {178},
pages = {105189},
year = {2025},
issn = {0968-090X},
doi = {https://doi.org/10.1016/j.trc.2025.105189},
url = {https://www.sciencedirect.com/science/article/pii/S0968090X25001937},
author = {Yantong Li and Shengjie Wang and Hairui Sun and Shanshan Zhou}
}

@article{Ref_Related_UAV5,
title = {Drone routing problem for shore-to-ship delivery services considering non-linear energy consumption},
journal = {Transportation Research Part B: Methodological},
volume = {206},
pages = {103410},
year = {2026},
issn = {0191-2615},
doi = {https://doi.org/10.1016/j.trb.2026.103410},
url = {https://www.sciencedirect.com/science/article/pii/S0191261526000226},
author = {Mengtong Wang and Shukai Chen and Qiang Meng}
}

@ARTICLE{Ref_Related_Off1,
  author={Chen, Xiangchun and Cao, Jiannong and Sahni, Yuvraj and Zhang, Mingjin and Liang, Zhixuan and Yang, Lei},
  journal={IEEE Transactions on Mobile Computing}, 
  title={Mobility-Aware Dependent Task Offloading in Edge Computing: A Digital Twin-Assisted Reinforcement Learning Approach}, 
  year={2025},
  volume={24},
  number={4},
  pages={2979-2994},
  doi={10.1109/TMC.2024.3506221}}

@ARTICLE{Ref_Related_Off2,
  author={Dong, Chongwu and Li, Weidong and Zhou, Zhi and Chen, Xu and Tian, Zhihong and Wen, Wushao},
  journal={IEEE Transactions on Mobile Computing}, 
  title={Delay-Sensitive Task Offloading With Edge Caching Through Martingale-Based Deep Reinforcement Learning}, 
  year={2025},
  volume={24},
  number={7},
  pages={6225-6242},
  doi={10.1109/TMC.2025.3540413}}

@ARTICLE{Ref_Related_Off3,
  author={Chen, Xiangchun and Cao, Jiannong and Cao, Rui and Sahni, Yuvraj and Zhang, Mingjin and Ji, Yusheng},
  journal={IEEE Transactions on Mobile Computing}, 
  title={Decentralized Task Offloading in Collaborative Edge Computing: A Digital Twin Assisted Multi-Agent Reinforcement Learning Approach}, 
  year={2026},
  volume={25},
  number={4},
  pages={4776-4790},
  doi={10.1109/TMC.2025.3628502}}

@ARTICLE{Ref_Related_Off4,
  author={Nabi, Ahmadun and Moh, Sangman},
  journal={IEEE Transactions on Mobile Computing}, 
  title={Joint Offloading Decision, User Association, and Resource Allocation in Hierarchical Aerial Computing: Collaboration of UAVs and HAP}, 
  year={2025},
  volume={24},
  number={8},
  pages={7267-7282},
  doi={10.1109/TMC.2025.3548668}}

@ARTICLE{Ref_Related_Off5,
  author={Gu, Huixian and Zhao, Liqiang and Han, Zhu and Chu, Xiaoli and Zheng, Gan and Liu, Jiaxin and Zhou, Guorong},
  journal={IEEE Transactions on Mobile Computing}, 
  title={Joint Task Offloading and Resource Allocation in Ultra-Dense Multi-Access Edge Computing: A Mean Field Learning Approach}, 
  year={2026},
  volume={25},
  number={3},
  pages={3598-3615},
  doi={10.1109/TMC.2025.3619077}}

@article{Ref_Related_Agent1,
title = {AI Agents and Agentic AI–navigating a plethora of concepts for future manufacturing},
journal = {Journal of Manufacturing Systems},
volume = {83},
pages = {126-133},
year = {2025},
issn = {0278-6125},
doi = {https://doi.org/10.1016/j.jmsy.2025.08.017},
url = {https://www.sciencedirect.com/science/article/pii/S027861252500216X},
author = {Yinwang Ren and Yangyang Liu and Tang Ji and Xun Xu}
}

@article{Ref_Related_Agent2,
title = {Agentic AI for smart manufacturing},
journal = {Manufacturing Letters},
volume = {46},
pages = {92-96},
year = {2025},
issn = {2213-8463},
doi = {https://doi.org/10.1016/j.mfglet.2025.10.013},
url = {https://www.sciencedirect.com/science/article/pii/S2213846325002883},
author = {Jay Lee and Hanqi Su}
}

@article{Ref_Related_Agent3,
title = {Application of retrieval-augmented generation for interactive industrial knowledge management via a large language model},
journal = {Computer Standards \& Interfaces},
volume = {94},
pages = {103995},
year = {2025},
issn = {0920-5489},
doi = {https://doi.org/10.1016/j.csi.2025.103995},
url = {https://www.sciencedirect.com/science/article/pii/S0920548925000248},
author = {Lun-Chi Chen and Mayuresh Sunil Pardeshi and Yi-Xiang Liao and Kai-Chih Pai}
}

@article{Ref_Related_Agent4,
title = {A4PS: Agentic AI-assisted advanced planning and scheduling with large language models for smart manufacturing},
journal = {Journal of Manufacturing Systems},
volume = {85},
pages = {207-226},
year = {2026},
issn = {0278-6125},
doi = {https://doi.org/10.1016/j.jmsy.2026.01.003},
url = {https://www.sciencedirect.com/science/article/pii/S0278612526000154},
author = {Mingxing Li and Qu Zhou and Wanshan Li and Ting Qu and Maolin Yang and Pingyu Jiang}
}

@article{Ref_Related_Agent5,
title = {Agentic Data Analysis for Intelligent Manufacturing: Benchmark-Driven Evaluation of Agentic vs. Direct LLM Approaches},
journal = {Procedia CIRP},
volume = {139},
pages = {280-285},
year = {2026},
note = {13th CIRP Global Web Conference},
issn = {2212-8271},
doi = {https://doi.org/10.1016/j.procir.2025.09.043},
url = {https://www.sciencedirect.com/science/article/pii/S2212827125010030},
author = {Nastaran Moradzadeh Farid and Alireza Taghizadeh and Sara Shafiee}
}

@article{Ref_CoT_Mfg3,
title = {MASC: Large language model-based multi-agent scheduling chain for flexible job shop scheduling problem},
journal = {Advanced Engineering Informatics},
volume = {67},
pages = {103527},
year = {2025},
issn = {1474-0346},
doi = {https://doi.org/10.1016/j.aei.2025.103527},
url = {https://www.sciencedirect.com/science/article/pii/S1474034625004203},
author = {Zelong Wang and Chenhui Wan and Jie Liu and Xi Zhang and Haifeng Wang and Youmin Hu and Zhongxu Hu}
}

@article{Ref_CoT_Mfg4,
title = {Large language model-empowered dynamic scheduling for intelligent hybrid flow shop using multi-agent deep reinforcement learning},
journal = {Advanced Engineering Informatics},
volume = {71},
pages = {104294},
year = {2026},
issn = {1474-0346},
doi = {https://doi.org/10.1016/j.aei.2025.104294},
url = {https://www.sciencedirect.com/science/article/pii/S1474034625011875},
author = {Wenbin Gu and Yushang Cao and Yuxin Li and Nuandong Li and Lei Wang and Na Tang and Minghai Yuan and Fengque Pei}
}

@article{Ref_IV_RAG3,
title = {Enhancing Retrieval-Augmented Generation with topic-enriched embeddings: A hybrid approach integrating traditional NLP techniques},
journal = {Natural Language Processing Journal},
volume = {14},
pages = {100200},
year = {2026},
issn = {2949-7191},
doi = {https://doi.org/10.1016/j.nlp.2026.100200},
url = {https://www.sciencedirect.com/science/article/pii/S294971912600004X},
author = {Rodrigo Kataishi}
}

@article{Ref_hyb2,
title = {Optimizing 3D trajectory and task offloading in collaborative UAV-Enabled mobile edge computing networks},
journal = {Computer Networks},
volume = {282},
pages = {112283},
year = {2026},
issn = {1389-1286},
doi = {https://doi.org/10.1016/j.comnet.2026.112283},
url = {https://www.sciencedirect.com/science/article/pii/S1389128626002951},
author = {Long Jiao and Ling Gao and Jie Zheng and Peiqing Yang and Zhaowei Zhang}
}

@ARTICLE{sun_UVA,
  author={Sun, Geng and Wu, Jiaxu and Sun, Zemin and He, Long and Wang, Jiacheng and Niyato, Dusit and Jamalipour, Abbas and Mao, Shiwen},
  journal={IEEE Transactions on Services Computing}, 
  title={J$\text{C}^{5}$A: Service Delay Minimization for Aerial MEC-Assisted Industrial Cyber-Physical Systems}, 
  year={2025},
  volume={18},
  number={5},
  pages={2976-2993},
  doi={10.1109/TSC.2025.3592419}}

@article{Ref_joint_LC1,
title = {Drone on-demand delivery routing problem considering order splitting and battery swapping},
journal = {Computers \& Industrial Engineering},
volume = {208},
pages = {111388},
year = {2025},
issn = {0360-8352},
doi = {https://doi.org/10.1016/j.cie.2025.111388},
url = {https://www.sciencedirect.com/science/article/pii/S0360835225005340},
author = {Shuxuan Li and Tianjun Liao and Guohua Wu and Yalin Wang and Ponnuthurai Nagaratnam Suganthan}
}

@ARTICLE{HanwenRAG1,
  author={Zhang, Hanwen and Zhang, Ruichen and Zhang, Wei and Niyato, Dusit and Wen, Yonggang and Miao, Chunyan},
  journal={IEEE Internet of Things Journal}, 
  title={Advancing Generative Artificial Intelligence and Large Language Models for Demand Side Management with Internet of Electric Vehicles}, 
  year={2026},
  volume={},
  number={},
  pages={1-1},
  doi={10.1109/JIOT.2026.3685302}}

@article{Ref_FC_s4,
title = {Towards evidence-aware retrieval-augmented generation via self-corrective chain-of-thought},
journal = {Information Processing \& Management},
volume = {63},
number = {2, Part A},
pages = {104369},
year = {2026},
issn = {0306-4573},
doi = {https://doi.org/10.1016/j.ipm.2025.104369},
url = {https://www.sciencedirect.com/science/article/pii/S0306457325003103},
author = {Yining Li and Wenjun Ke and Jiajun Liu and Peng Wang and Jianghan Liu and Yao He}
}

@article{R2_Q2_Ref1,
title = {Dynamic-aware deep reinforcement learning for heterogeneous fleet-based capacitated electric vehicle routing problems},
journal = {Applied Energy},
volume = {413},
pages = {127757},
year = {2026},
issn = {0306-2619},
doi = {https://doi.org/10.1016/j.apenergy.2026.127757},
url = {https://www.sciencedirect.com/science/article/pii/S0306261926004095},
author = {Qingshu Guan and Hui Cao and Junkai Tan and Bei Ou and Yanbo Wang and Xingyu Yan and Badong Chen}
}

@article{R2_Q2_Ref2,
title = {Heterogeneous attention-driven deep reinforcement learning for solving EVRPs with pickup-delivery and time windows},
journal = {Applied Energy},
volume = {407},
pages = {127355},
year = {2026},
issn = {0306-2619},
doi = {https://doi.org/10.1016/j.apenergy.2026.127355},
url = {https://www.sciencedirect.com/science/article/pii/S0306261926000073},
author = {Qingshu Guan and Hui Cao and Tiansen Niu and Lixin Jia and Dapeng Yan and Badong Chen}
}

@article{Typical_CoT1,
title = {Unleashing the potential of prompt engineering for large language models},
journal = {Patterns},
volume = {6},
number = {6},
pages = {101260},
year = {2025},
issn = {2666-3899},
doi = {https://doi.org/10.1016/j.patter.2025.101260},
url = {https://www.sciencedirect.com/science/article/pii/S2666389925001084},
author = {Banghao Chen and Zhaofeng Zhang and Nicolas Langrené and Shengxin Zhu}
}

@article{Typical_CoT2,
title = {Key-concept thinking prompting for improved reasoning in large language models},
journal = {Neurocomputing},
volume = {656},
pages = {130986},
year = {2025},
issn = {0925-2312},
doi = {https://doi.org/10.1016/j.neucom.2025.130986},
url = {https://www.sciencedirect.com/science/article/pii/S0925231225016583},
author = {Minghua Tang and Chen Bian and Liming Yang and Xueling Zhong}
}

@article{Typical_CoT3,
title = {Improving intermediate reasoning in zero-shot chain-of-thought for large language models with filter supervisor-self correction},
journal = {Neurocomputing},
volume = {620},
pages = {129219},
year = {2025},
issn = {0925-2312},
doi = {https://doi.org/10.1016/j.neucom.2024.129219},
url = {https://www.sciencedirect.com/science/article/pii/S0925231224019908},
author = {Jun Sun and Yiteng Pan and Xiaohu Yan}
}

@inproceedings{ref_quant_evalu1,
    title = "{RAGA}s: Automated Evaluation of Retrieval Augmented Generation",
    author = "Es, Shahul  and
      James, Jithin  and
      Espinosa Anke, Luis  and
      Schockaert, Steven",
    editor = "Aletras, Nikolaos  and
      De Clercq, Orphee",
    booktitle = "Proceedings of the 18th Conference of the European Chapter of the Association for Computational Linguistics: System Demonstrations",
    month = mar,
    year = "2024",
    address = "St. Julians, Malta",
    publisher = "Association for Computational Linguistics",
    url = "https://aclanthology.org/2024.eacl-demo.16/",
    doi = "10.18653/v1/2024.eacl-demo.16",
    pages = "150--158"
}

@ARTICLE{ref_quant_evalu2,
  author={Elkiran, Harun and Rasheed, Jawad},
  journal={IEEE Access}, 
  title={EvaRAG: Evaluating Advanced RAG Techniques With Indexing and Distance Metrics}, 
  year={2025},
  volume={13},
  number={},
  pages={215724-215747},
  doi={10.1109/ACCESS.2025.3646665}}

@article{ref_quant_evalu3,
title = {HaluCheck: Explainable and verifiable automation for detecting hallucinations in LLM responses},
journal = {Expert Systems with Applications},
volume = {272},
pages = {126712},
year = {2025},
issn = {0957-4174},
doi = {https://doi.org/10.1016/j.eswa.2025.126712},
url = {https://www.sciencedirect.com/science/article/pii/S0957417425003343},
author = {Sangwoo Heo and Sungwook Son and Hyunwoo Park}
}

@article{ref_quant_evalu4,
title = {Knowledge Graphs, Large Language Models, and Hallucinations: An NLP Perspective},
journal = {Journal of Web Semantics},
volume = {85},
pages = {100844},
year = {2025},
issn = {1570-8268},
doi = {https://doi.org/10.1016/j.websem.2024.100844},
url = {https://www.sciencedirect.com/science/article/pii/S1570826824000301},
author = {Ernests Lavrinovics and Russa Biswas and Johannes Bjerva and Katja Hose}
}

@article{Ref_DRL_state1,
title = {Action decoupled SAC reinforcement learning with discrete-continuous hybrid action spaces},
journal = {Neurocomputing},
volume = {537},
pages = {141-151},
year = {2023},
issn = {0925-2312},
doi = {https://doi.org/10.1016/j.neucom.2023.03.054},
url = {https://www.sciencedirect.com/science/article/pii/S0925231223003028},
author = {Yahao Xu and Yiran Wei and Keyang Jiang and Li Chen and Di Wang and Hongbin Deng}
}

@article{Ref_DRL_state2,
title = {The application of machine learning based energy management strategy in multi-mode plug-in hybrid electric vehicle, part I: Twin Delayed Deep Deterministic Policy Gradient algorithm design for hybrid mode},
journal = {Energy},
volume = {262},
pages = {125084},
year = {2023},
issn = {0360-5442},
doi = {https://doi.org/10.1016/j.energy.2022.125084},
url = {https://www.sciencedirect.com/science/article/pii/S036054422201979X},
author = {Changcheng Wu and Jiageng Ruan and Hanghang Cui and Bin Zhang and Tongyang Li and Kaixuan Zhang}
}

@ARTICLE{Ref_DRL_state3,
  author={Mao, Xiao and Wu, Guohua and Fan, Mingfeng and Cao, Zhiguang and Pedrycz, Witold},
  journal={IEEE Transactions on Automation Science and Engineering}, 
  title={DL-DRL: A Double-Level Deep Reinforcement Learning Approach for Large-Scale Task Scheduling of Multi-UAV}, 
  year={2025},
  volume={22},
  number={},
  pages={1028-1044},
  doi={10.1109/TASE.2024.3358894}}

@article{Ref_DRL_design1,
title = {An O-MAPPO scheme for joint computation offloading and resources allocation in UAV assisted MEC systems},
journal = {Computer Communications},
volume = {208},
pages = {190-199},
year = {2023},
issn = {0140-3664},
doi = {https://doi.org/10.1016/j.comcom.2023.06.008},
url = {https://www.sciencedirect.com/science/article/pii/S0140366423002037},
author = {Ming Cheng and Canlin Zhu and Min Lin and Jun-Bo Wang and Wei-Ping Zhu}
}

@article{Ref_DRL_design2,
title = {A multi-UAV assisted task offloading and path optimization for mobile edge computing via multi-agent deep reinforcement learning},
journal = {Journal of Network and Computer Applications},
volume = {229},
pages = {103919},
year = {2024},
issn = {1084-8045},
doi = {https://doi.org/10.1016/j.jnca.2024.103919},
url = {https://www.sciencedirect.com/science/article/pii/S1084804524000961},
author = {Tao Ju and Linjuan Li and Shuai Liu and Yu Zhang}
}

@article{CP1_GraphThought,
  title={GraphThought: Graph Combinatorial Optimization with Thought Generation},
  author={Zixiao Huang and Lifeng Guo and Junjie Sheng and Haosheng Chen and Wenhao Li and Bo Jin and Changhong Lu and Xiangfeng Wang},
  journal={arXiv:2502.11607},
  year={2025},
  url={https://arxiv.org/abs/2502.11607}
}

@article{JSP3_LLM,
  title={LLMs can Schedule},
  author={Henrik Abgaryan and Ararat Harutyunyan and Tristan Cazenave},
  journal={arXiv:2408.06993},
  year={2024},
  url={https://arxiv.org/abs/2408.06993}
}

@article{JSP2_LLM,
  title={A Large Language Model-based multi-agent manufacturing system for intelligent shopfloor},
  author={Zhen Zhao and Dunbing Tang and Haihua Zhu and Zequn Zhang and Kai Chen and Changchun Liu and Yuchen Ji},
  journal={arXiv:2405.16887},
  year={2024},
  url={https://arxiv.org/abs/2405.16887}
}

@article{Ref_FC_s1,
  title={Collab-RAG: Boosting Retrieval-Augmented Generation for Complex Question Answering via White-Box and Black-Box LLM Collaboration},
  author={Ran Xu and Wenqi Shi and Yuchen Zhuang and Yue Yu and Joyce C. Ho and Haoyu Wang and Carl Yang},
  journal={arXiv:2504.04915},
  year={2025},
  url={https://arxiv.org/abs/2504.04915}
}

@article{Ref_FC_s2,
  title={MA-RAG: Multi-Agent Retrieval-Augmented Generation via Collaborative Chain-of-Thought Reasoning},
  author={Thang Nguyen and Peter Chin and Yu-Wing Tai},
  journal={arXiv:2505.20096},
  year={2025},
  url={https://arxiv.org/abs/2505.20096}
}

@article{Ref_FC_s3,
  title={MARS: toward more efficient multi-agent collaboration for LLM reasoning},
  author={Xiao Wang and Jia Wang and Yijie Wang and Pengtao Dang and Sha Cao and Chi Zhang},
  journal={arXiv:2509.20502},
  year={2026},
  url={https://arxiv.org/abs/2509.20502}
}

@article{Ref_F_CoT1,
  title={Towards Reasoning Era: A Survey of Long Chain-of-Thought for Reasoning Large Language Models},
  author={Qiguang Chen and Libo Qin and Jinhao Liu and Dengyun Peng and Jiannan Guan and Peng Wang and Mengkang Hu and Yuhang Zhou and Te Gao and Wanxiang Che},
  journal={arXiv:2503.09567},
  year={2025},
  url={https://arxiv.org/abs/2503.09567}
}

@article{Ref_F_CoT2,
  title={Interactive Reasoning: Visualizing and Controlling Chain-of-Thought Reasoning in Large Language Models},
  author={Rock Yuren Pang and K. J. Kevin Feng and Shangbin Feng and Chu Li and Weijia Shi and Yulia Tsvetkov and Jeffrey Heer and Katharina Reinecke},
  journal={arXiv:2506.23678},
  year={2025},
  url={https://arxiv.org/abs/2506.23678}
}

@article{Ref_F_CoT3,
  title={Zero-Shot Verification-guided Chain of Thoughts},
  author={Jishnu Ray Chowdhury and Cornelia Caragea},
  journal={arXiv:2501.13122},
  year={2025},
  url={https://arxiv.org/abs/2501.13122}
}

@article{Ref_F_CoT4,
  title={Evaluating Chain-of-Thought Reasoning through Reusability and Verifiability},
  author={Shashank Aggarwal and Ram Vikas Mishra and Amit Awekar},
  journal={arXiv:2602.17544},
  year={2026},
  url={https://arxiv.org/abs/2602.17544}
}

@article{UseCase2_R12_update,
   author = {Khosiawan, Yohanes and Park, Youngsoo and Moon, Ilkyeong and Nilakantan, Janardhanan Mukund and Nielsen, Izabela},
   title = {Task scheduling system for UAV operations in indoor environment},
   journal = {Neural Computing and Applications},
   volume = {31},
   number = {9},
   pages = {5431-5459},
   ISSN = {1433-3058},
   DOI = {10.1007/s00521-018-3373-9},
   url = {https://doi.org/10.1007/s00521-018-3373-9},
   year = {2019},
   type = {Journal Article}
}

@book{Cosine_Similarity_update,
  author = {Manning, Christopher D. and Raghavan, Prabhakar and Sch{\"u}tze, Hinrich},
  title = {Introduction to Information Retrieval},
  publisher = {Cambridge University Press},
  address = {Cambridge, UK},
  year = {2008},
  isbn = {978-0-521-86571-5}
}

@misc{text_embedding_ada_002,
  author = {OpenAI},     
  title = "text-embedding-ada-002",
  howpublished = {\url{https://developers.openai.com/api/docs/models/text-embedding-ada-002}},
  year = "2026",
  note = {Accessed: 03 Apr 2026}
}

@misc{gpt_5dot6_Luna,
  author = {OpenAI},     
  title = "{GPT-5.6 Luna}",
  howpublished = {\url{https://developers.openai.com/api/docs/models/gpt-5.6-luna}},
  year = "2026",
  note = {Accessed: 14 Jul 2026}
}

@misc{Chroma,
  author = {Chroma},     
  title = "What Chroma Offers",
  howpublished = {\url{https://docs.trychroma.com/docs/overview/introduction}},
  year = "2026",
  note = {Accessed: 03 Apr 2026}
}

@misc{LangGraph,
  author = {LangChain},     
  title = "LangGraph overview",
  howpublished = {\url{https://docs.langchain.com/oss/python/langgraph/overview}},
  year = "2026",
  note = {Accessed: 03 Apr 2026}
}

@misc{LangChain,
  author = {LangChain},     
  title = "LangChain overview",
  howpublished = {\url{https://docs.langchain.com/oss/javascript/langchain/overview}},
  year = "2026",
  note = {Accessed: 30 Jun 2026}
}



\end{document}